\documentclass[preprint, 12pt, times, number]{elsarticle}
\usepackage{amsmath, amssymb, amsfonts}
\usepackage{graphicx}
\usepackage{booktabs}
\usepackage[table]{xcolor}
\usepackage{multirow}
\usepackage{subcaption}
\usepackage{longtable}
\usepackage{array}
\usepackage{float}
\usepackage{enumitem}
\usepackage{pifont}
\usepackage{url}
\newcommand{\cmark}{\textcolor{green!50!black}{\ding{51}}}
\newcommand{\xmark}{\textcolor{red}{\ding{55}}}
\usepackage{natbib}
\usepackage{hyperref}
\definecolor{linkpurple}{RGB}{128, 0, 128}
\journal{Computers and Electronics in Agriculture}
\begin{document}
\begin{frontmatter}
\title{\raisebox{-0.3\height}{\includegraphics[height=1.8em]{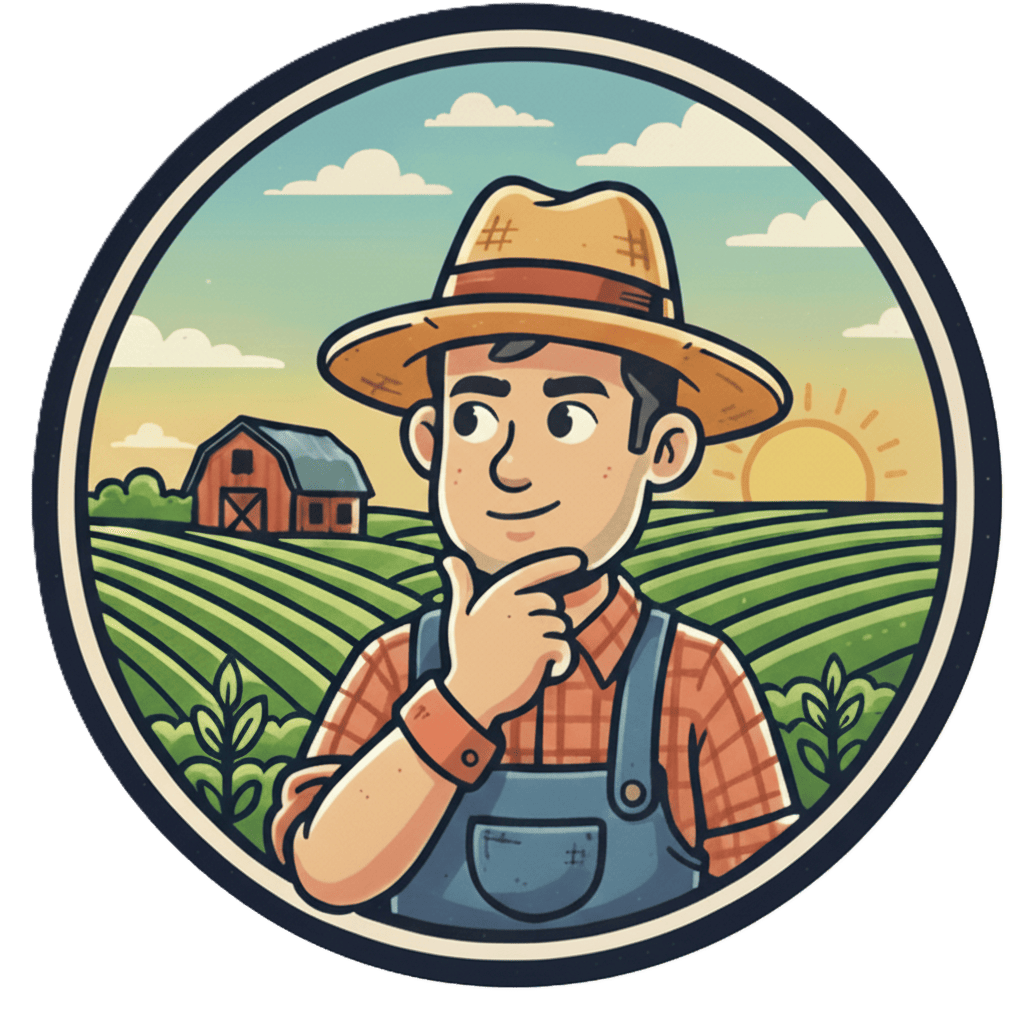}} AgriChat: A Multimodal Large Language Model for Agriculture Image Understanding}

\author[label1]{Abderrahmene Boudiaf\corref{cor1}}
\ead{100058322@ku.ac.ae}

\author[label1]{Irfan Hussain}
\author[label1]{Sajid Javed}

\address[label1]{Department of Computer Science, Khalifa University of Science and Technology,
Abu Dhabi, United Arab Emirates}

\cortext[cor1]{Corresponding author}

\begin{abstract}
The deployment of Multimodal Large Language Models (MLLMs) in agriculture is currently stalled by a critical trade-off: the existing literature lacks the large-scale agricultural datasets required for robust model development and evaluation, while current state-of-the-art models lack the verified domain expertise necessary to reason across diverse taxonomies. To address these challenges, we propose the \textit{Vision-to-Verified-Knowledge (V2VK)} pipeline, a novel generative AI-driven annotation framework that integrates visual captioning with web-augmented scientific retrieval to autonomously generate the \textit{AgriMM} benchmark, effectively eliminating biological hallucinations by grounding training data in verified phytopathological literature. The AgriMM benchmark contains over 3,000 agricultural classes and more than 607k VQAs spanning multiple tasks, including fine-grained plant species identification, plant disease symptom recognition, crop counting, and ripeness assessment. Leveraging this verifiable data, we present \textit{AgriChat}, a specialized MLLM that presents broad knowledge across thousands of agricultural classes and provides detailed agricultural assessments with extensive explanations. Extensive evaluation across diverse tasks, datasets, and evaluation conditions reveals both the capabilities and limitations of current agricultural MLLMs, while demonstrating AgriChat's superior performance over other open-source models, including internal and external benchmarks. The results validate that preserving visual detail combined with web-verified knowledge constitutes a reliable pathway toward robust and trustworthy agricultural AI. The code and dataset are publicly available at \textcolor{linkpurple}{\href{https://github.com/boudiafA/AgriChat}{\texttt{https://github.com/boudiafA/AgriChat}}}.
\end{abstract}

\begin{keyword}
Multimodal Large Language Models \sep Foundation models \sep Agricultural \sep Plant Identification \sep Plant Disease Analysis \sep Crop counting.
\end{keyword}

\end{frontmatter}

\section{Introduction}
\label{sec:intro}
\begin{figure}[htbp]
\centering
\includegraphics[width=\linewidth]{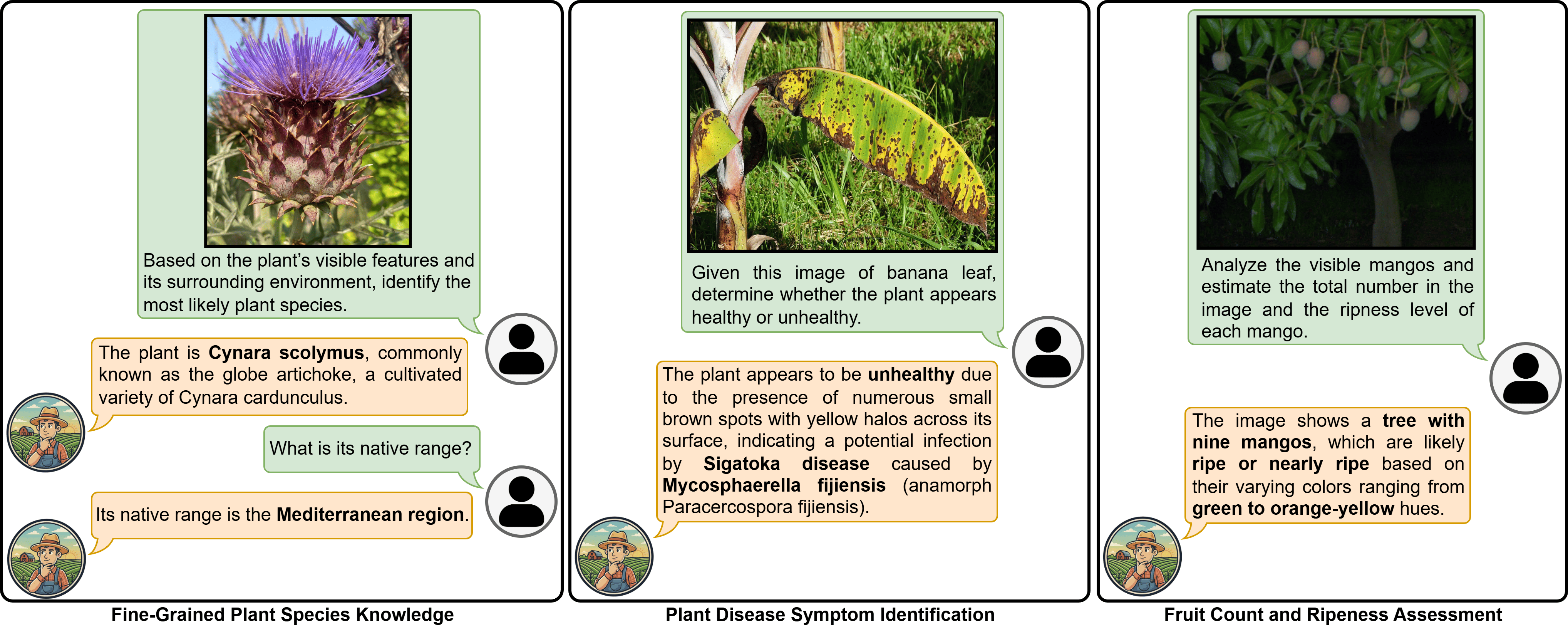}
\caption{Illustrative examples of AgriChat's conversational diagnostic capabilities across three core agricultural tasks. \textbf{Left:} Fine-grained plant species identification with follow-up knowledge queries. \textbf{Center:} Plant disease symptom recognition, correctly diagnosing Sigatoka disease from visible leaf lesions. \textbf{Right:} Fruit counting and ripeness assessment from a single field image. These examples highlight AgriChat's ability to support interactive, expert-level agricultural reasoning beyond simple classification.}
\label{fig:agrichat_qualitative_intro}
\end{figure}

\begin{figure}[htbp]
\centering
\includegraphics[width=0.95\linewidth]{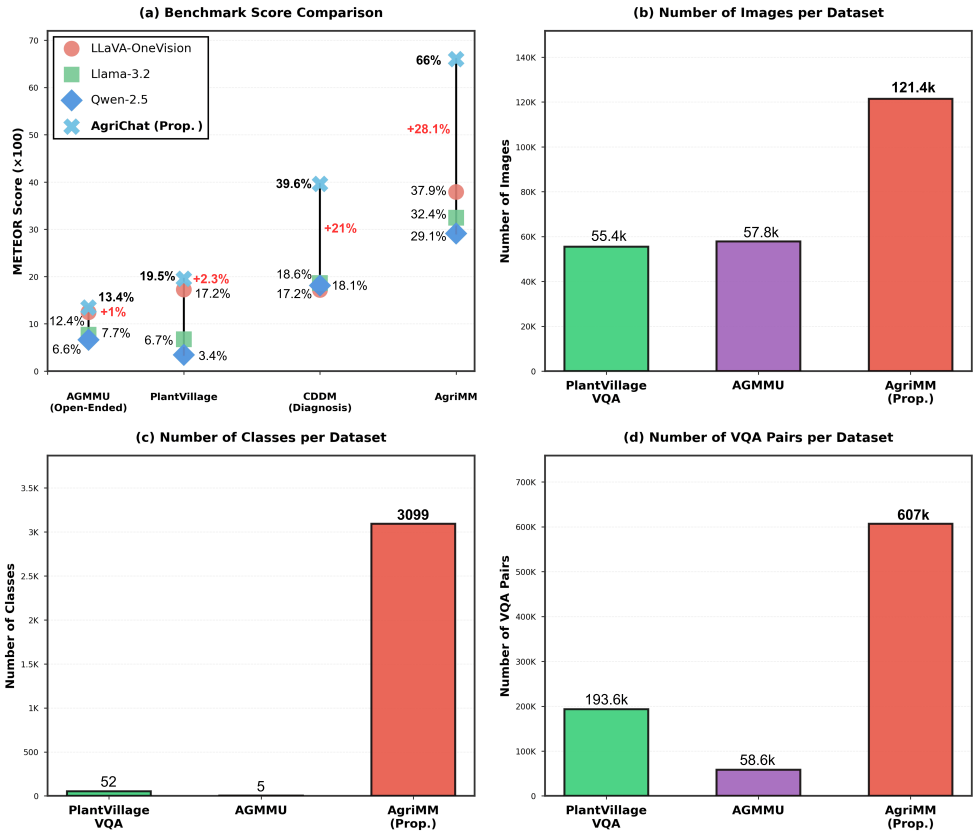}
\caption{\textbf{(a)} Comparison between existing foundation models (Llama-3.2 \cite{llama3-2024}, LLava-OneVision \cite{llava-onevision2024}, Qwen-2.5 \cite{qwen2024}) and our proposed agriculture-focused \textbf{AgriChat} model which achieved superior performance across four agriculture benchmarks. In comparison to existing agriculture benchmark datasets (PlantVillageVQA \cite{plantvillagevqa2024}, AGMMU \cite{AGMMU}), our proposed large-scale benchmark \textbf{AgriMM} exceeds the state-of-the-art benchmarks in terms of \textbf{(b)} number of images, \textbf{(c)} number of classes, and \textbf{(d)} number of VQA pairs.}
\label{fig:}
\end{figure}

Agriculture remains the cornerstone of global economic stability and food security, sustaining approximately 80\% of the world’s rural poor and serving as the primary instrument for poverty alleviation \cite{worldbank2024, ifad2024}. As the global population is projected to approach 10 billion by 2050, the agriculture sector faces mounting pressure to increase production by an estimated 70\% while simultaneously navigating the existential threats of climate change, soil degradation, and diminishing arable land \cite{fao2024}. Artificial Intelligence (AI) has emerged as a transformative force, shifting traditional farming paradigms toward data-driven precision agriculture \cite{adewusi2024}. Recent studies indicate that AI-integrated systems ranging from predictive analytics for yield optimization to computer vision for early disease detection enhance crop productivity by up to 20\% while reducing water and fertilizer inputs by nearly 30\% \cite{zhang2025, kumar2024}.
Consequently, the integration of AI tools is no longer merely advantageous but essential for fostering resilient agricultural systems capable of meeting future nutritional demands within planetary boundaries \cite{mercatoris2024}.

In the literature, the Convolutional Neural Networks (CNNs) have driven much of this progress, they typically function as ``black boxes,'' providing classification labels without the explanatory reasoning required for effective decision support \cite{nieradzik2024challenging}.
This limitation has catalyzed interest in Multimodal Large Language Models (MLLMs), which offer the ability to reason over visual inputs and provide interactive diagnostics \cite{agrogpt2024}. However, the efficacy of MLLMs is strictly bound by the quality of the data upon which they are trained \cite{yin2025}. Existing agricultural Visual Question Answering (VQA) datasets suffer from a fundamental trade-off: they either rely on expensive human annotation that limits scale and taxonomic diversity (e.g., PlantVillage with only 38 diseases \cite{hughes2015plantvillage}, CDDM with 60 disease types biased toward commodity crops \cite{cddm}), or they employ synthetic generation using frozen LLMs that may introduce incorrect hallucinated information without real-time verification. Historical expert logs like AgMMU, while factually accurate at creation, suffer from knowledge staleness as agricultural best practices, pesticide regulations, and pathogen management protocols continuously evolve \cite{AGMMU}.

To address this gap, we introduce the Agriculture Multi-Modal dataset (AgriMM), which resolves both limitations simultaneously through a novel approach to data curation. Unlike previous benchmarks, AgriMM was created by systematically aggregating and filtering 63 distinct agricultural datasets to ensure broad taxonomic coverage, spanning more than 3,000 classes and 121,000 images, containing questions related to fine-grained species identification, disease recognition, and crop counting tasks. The core novelty lies in our Vision-to-Verified-Knowledge (V2VK) pipeline, a three-stage synthesis process designed to achieve diverse, accurate, and verifiably grounded knowledge. First, we employ Gemma 3~ \cite{gemma2024} to generate detailed visual captions for each agricultural image, capturing key visual attributes such as plant morphology, growth stage, visible symptoms of disease when present, and other agronomically relevant features. Second, utilizing a Retrieval-Augmented Generation (RAG) framework, we leverage Gemini 3 Pro~ \cite{gemini2024} with real-time web access to produce a comprehensive class-level description covering taxonomy, biology, and known diseases, drawing from authoritative sources to filter out hallucinations and ensure factual accuracy. Finally, LLaMA 3.1~ \cite{llama3-2024} synthesizes both the image caption and the class description into 607,000 diverse instruction-tuning question-answer pairs. This rigorous curation ensures that the ground truth textual descriptions are not merely linguistically plausible, but scientifically current.

Despite the availability of such data, current model architectures present a secondary bottleneck. Existing agricultural MLLMs including Agri-LLaVA \cite{agri-llava2024}, AgroGPT \cite{agrogpt2024}, and LLaVA-PlantDiag \cite{llava-plantdiag2024} universally suffer from limited and narrow fine-tuning data: they are trained on small-scale, task-specific corpora that cover only a handful of crops and diseases, restricting their ability to generalize across the vast taxonomic and pathological diversity encountered in real-world agricultural settings \cite{agrogpt2024, agri-llava2024}. For instance, models fine-tuned on fewer than 200 classes inevitably fail when confronted with underrepresented regional cultivars, emerging pathogens, or multi-task scenarios requiring simultaneous species identification, disease diagnosis, and yield estimation \cite{global-crop-diversity2023}. In this work, we propose \textbf{AgriChat}, an MLLM framework that fundamentally overcomes this data bottleneck. AgriChat is fine-tuned on our large-scale \textbf{AgriMM} dataset comprising over 121k images and over 607k QA pairs spanning more than 3,000 classes, representing the widest and most diverse agricultural fine-tuning corpus to date.

To effectively adapt the base architecture to the agricultural domain while preserving its general-purpose reasoning capabilities, AgriChat employs a parameter-efficient fine-tuning strategy using Low-Rank Adaptation (LoRA) \cite{hu2022lora}. Lightweight trainable adapters are injected into both the vision encoder and the language model decoder, while all pre-trained weights remain frozen. The model undergoes supervised fine-tuning on the AgriMM corpus, optimizing an autoregressive objective with role-aware masking. This training paradigm allows AgriChat to acquire expert-level agricultural diagnostic knowledge, covering fine-grained species identification, disease recognition, and crop counting, while retaining the vast world knowledge inherent in the base model, all within a computationally efficient footprint (inference in $\sim$2.3 seconds on consumer hardware).

We validate the efficacy of our approach through rigorous experiments across four diverse benchmarks, including AgriMM, CDDM \cite{cddm}, PlantVillageVQA \cite{plantvillagevqa2024}, and AGMMU \cite{AGMMU}. Our results demonstrate that AgriChat not only achieves state-of-the-art performance on in-domain tasks but also exhibits superior zero-shot generalization to unseen datasets compared to larger open-source models. The main contributions of this paper are summarized as follows:
\begin{enumerate}
    \item \textbf{Agriculture Multimodal Dataset (AgriMM):} We introduce a high-quality instruction-tuning dataset comprising over 121k images and 607k expert-aligned QA pairs. By employing a web-search augmented generation pipeline, we solve the issue of knowledge staleness and hallucination, ensuring taxonomic diversity across 3,000+ agricultural classes.
    
    \item \textbf{AgriChat MLLM:} We introduce the first multimodal large language model (MLLM) purpose-built for agriculture and fine-tuned on the widest and most diverse range of agricultural species to date, enabling state-of-the-art performance on fine-grained species identification, disease classification and crop counting tasks.

    
    \item \textbf{Comprehensive Benchmarking:} Extensive evaluation on four agriculture benchmarks demonstrates that our proposed AgriChat achieves superior performance in both in-domain tasks and zero-shot generalization, outperforming state-of-the-art generalist baselines.
\end{enumerate}

The rest of this paper is organized as follows: Section~\ref{sec:related_work} reviews the evolution of agricultural AI and analyzes the limitations of current VQA datasets. Section~\ref{sec:methodology} details the construction of the AgriMM dataset through our Vision-to-Verified-Knowledge pipeline and presents the AgriChat architecture. Section~\ref{sec:results} presents the experimental setup, quantitative comparisons, and qualitative analysis. Finally, Section~\ref{sec:conclusion} concludes the paper and outlines future directions.

\section{Related Work} \label{sec:related_work}

\subsection{Existing Agriculture Datasets and Tasks}
The development of robust agricultural AI has been historically constrained by the availability and quality of training data. Early benchmarks like PlantVillage \cite{hughes2015open} provided a foundational resource with 54,306 images covering 38 disease classes. However, its reliance on laboratory-controlled imaging limits its applicability to complex field environments. More recent efforts have attempted to scale up visual diversity. The CDDM (Crop Disease Diagnosis Multimodal) benchmark \cite{cddm} expanded the scope to 137,000 images and 1 million QA pairs. Yet, as noted in recent critiques, its taxonomic scope remains constrained to 60 disease types across 16 major crop categories, reflecting a fundamental bias toward economically significant commodity crops (e.g., maize, rice, wheat) while leaving regional heirlooms and emerging pathogens underrepresented \cite{global-crop-diversity2023}. To address the prohibitive cost of manual annotation, researchers have turned to synthetic generation and historical log mining. AgroInstruct \cite{agrogpt2024} utilized diverse vision-only datasets to synthesize 70,000 instructions. While scalable, this approach relies on frozen language models that suffer from hallucination, generating plausible but factually incorrect statements about disease etiology or management protocols that were never true \cite{hallucination-llm2023}. Conversely, the AGMMU benchmark \cite{AGMMU} curated 57,079 expert-farmer dialogues from historical extension logs. While factually accurate at the time of creation, these datasets suffer from ``knowledge staleness'': agricultural best practices evolve continuously as pesticide regulations change and new pathogen strains appear. Expert dialogues from 2020 may no longer reflect current recommendations in 2026 \cite{fao-plantprotection2023}. Furthermore, a critical barrier to advancing agricultural AI is the \textit{reproducibility crisis}. As detailed in Table~\ref{tab:agri_vqa_datasets}, the largest corpus, Agri-3M-VL, remains unpublished. This prevents independent verification of reported performance and hinders regional adaptation, where researchers might otherwise fine-tune models for local crops. To bridge these gaps, we introduce AgriMM. Unlike prior datasets, AgriMM is a Multi-Source dataset consolidating 63 data sources into 121,425 images and 607,125 QA pairs. It integrates three critical tasks: fine-grained species identification, disease diagnosis, and crop counting spanning 3,099 classes. 

\subsection{Vision Models in Agriculture}
Deep learning in agriculture initially focused on pure visual classification using Convolutional Neural Networks (CNNs). Architectures such as ResNet \cite{he2016deep}, DenseNet \cite{huang2017densely}, and EfficientNet \cite{tan2019efficientnet} established the dominant paradigm, achieving impressive accuracy rates exceeding 95\% on controlled datasets. Recent advancements have introduced hybrid CNN-Transformer architectures to capture global context. Notable examples include LGNet \cite{lgnet2024}, which employs dual-branch adaptive feature fusion, and ST-CFI \cite{stcfi2025}, which implements bidirectional fusion of CNN spatial features with Swin Transformer \cite{liu2021swin} tokens. However, despite their superior classification performance, all such systems fundamentally operate as \textit{black-box} discriminative models. They accept an image as input and return only a categorical label (e.g., ``Tomato Late Blight'') without explanatory reasoning, contextual information, or interactive diagnostic capabilities. This lack of interpretability limits their utility in decision support systems, where farmers require not just a diagnosis, but an understanding of the cause and appropriate management strategies.

\subsection{Vision Language Models in Agriculture}
To overcome the limitations of closed-set classification, researchers began exploring Vision-Language Models (VLMs) that align image features with natural language representations. Early attempts such as PlantVillageVQA \cite{plantvillagevqa2024} introduced question-answer pairs but suffered from rigid templates (e.g., ``What disease does this plant have?''), resulting in limited linguistic diversity and shallow reasoning depth. More recent work has leveraged large-scale pretraining. ALive \cite{agriclip2025} provides approximately 600,000 image-text pairs covering crops and livestock to train AgriCLIP. By aligning visual encoders with agricultural text, AgriCLIP demonstrated significant gains (9.07\%) in zero-shot classification over standard CLIP models. However, these models are primarily designed for retrieval and matching tasks. While they can associate an image of a leaf with the text ``boron deficiency,'' they lack the generative capacity to engage in multi-turn dialogue or provide nuanced reasoning about symptom progression required for interactive agricultural decision support.

\subsection{MLLMs in Agriculture}
The most recent paradigm shift involves Multimodal Large Language Models (MLLMs), which integrate visual encoders with Large Language Models (LLMs) to enable conversational diagnostics. \textbf{Agri-LLaVA} \cite{agri-llava2024} pioneered knowledge-infused training by incorporating agricultural domain expertise into the LLaVA architecture. Similarly, \textbf{AgroGPT} \cite{agrogpt2024} and \textbf{LLaVA-PlantDiag} \cite{llava-plantdiag2024} have pushed the boundaries of interactive diagnosis, with the latter achieving BLEU-4 scores of 48.7\% on conversational diagnostic tasks. However, a critical limitation shared by all existing agricultural MLLMs is their reliance on narrow, small-scale fine-tuning datasets that cover only a limited number of crops and disease classes. This data bottleneck constrains their capacity to generalize across the vast taxonomic and pathological diversity encountered in real-world agricultural deployments, particularly for underrepresented regional cultivars, emerging pathogens, and multi-task scenarios requiring simultaneous species identification, disease diagnosis, and quantitative yield estimation. In this work, we propose AgriChat to address this data bottleneck. As illustrated in Figure~\ref{fig:agrichat_qualitative_intro}, AgriChat supports interactive, expert-level diagnostic reasoning across a diverse range of agricultural tasks from fine-grained plant species identification and multi-turn knowledge queries, to pathogen-level disease diagnosis and fruit counting with ripeness assessment. AgriChat is fine-tuned via Low-Rank Adaptation (LoRA)~ \cite{hu2022lora} on our proposed AgriMM dataset. This parameter-efficient adaptation strategy injects domain-specific agricultural knowledge into both the vision encoder and the language backbone while preserving the general-purpose reasoning capabilities of the base model.

\section{Methodology}
\label{sec:methodology}

\subsection{AgriMM Dataset}
\label{subsec:agrimm_dataset}

To address the scarcity of high-quality, verifiable agricultural data, we introduce AgriMM, a large-scale visual benchmark that represents a robust consolidation of diverse agricultural environments. AgriMM aggregates 63 source datasets, comprising 1 fine-grained taxonomic dataset (iNatAg subset \cite{agml}), 33 counting/detection datasets, and 29 disease classification datasets (see ~\ref{app:datasets} for complete source list), to create a total of 121,425 images annotated with 607,125 Visual Question Answering (VQA) pairs, covering over 3,000 distinct classes. The dataset architecture is built upon three complementary components, ensuring coverage across species-level identification, pathological diagnosis, and quantitative crop monitoring. Unlike existing datasets that often suffer from label noise and taxonomic inconsistencies, AgriMM is created through a rigorous curation pipeline that balances broad taxonomic coverage with specific, high-value agronomic tasks. To contextualize the contribution of AgriMM, Table~\ref{tab:agri_vqa_datasets} compares it against prominent agricultural VQA benchmarks. Existing datasets such as CDDM \cite{cddm} and PlantVillageVQA \cite{plantvillagevqa2024} are limited by narrow task definitions (lacking counting or fine-grained classification) or rely on static, potentially outdated knowledge bases. While Agri-3M-VL \cite{agri-3m-vl} offers scale, it remains unavailable to the research community. AgriMM addresses these limitations by offering the first publicly available, multi-source benchmark that integrates fine-grained taxonomy, counting capabilities, and a dynamic Web-RAG verification pipeline.

\begin{table*}[ht]
\centering
\caption{Comparison of Agricultural VQA Datasets. AgriMM stands out by integrating 63 data sources and utilizing a Web-RAG pipeline to ensure knowledge is factual and up-to-date, unlike static or hallucination-prone alternatives.}
\label{tab:agri_vqa_datasets}
\resizebox{\textwidth}{!}{%
\begin{tabular}{l rr c c c c c c}
\toprule
\textbf{Dataset Name} & \textbf{\# Images} & \textbf{\# QA Pairs} & \textbf{\# Classes} & \textbf{Multi-Source} & \textbf{Fine-Grained} & \textbf{Counting} & \textbf{Web-Verified} & \textbf{Available} \\ 
\midrule
\textbf{CDDM \cite{cddm}} & 137,000 & 1,000,000 & 76 & \xmark & \xmark & \xmark & \xmark & \cmark \\ 
\textbf{PlantVillageVQA \cite{plantvillagevqa2024}} & 55,448 & 193,609 & 52 & \xmark & \xmark & \xmark & \cmark$^{\ast}$ & \cmark \\ 
\textbf{Agri-3M-VL \cite{agri-3m-vl}} & $\sim$1.0M & $\sim$3.0M & 42,253 & \cmark & \cmark & \xmark & \xmark & \xmark \\ 
\textbf{AgroInstruct \cite{agrogpt2024}} & 108,701 & 70,000 & 202 & \cmark & \xmark & \xmark & \xmark & \xmark \\ 
\textbf{AGMMU \cite{AGMMU}} & 57,079 & 58,571 & 5 & \xmark & \xmark & \xmark & \cmark$^{\ast}$ & \cmark \\ 
\midrule
\rowcolor{gray!10} 
\textbf{AgriMM (Ours)} & \textbf{121,425} & \textbf{607,125} & \textbf{3,099} & \textbf{\cmark} & \textbf{\cmark} & \textbf{\cmark} & \textbf{\cmark} & \textbf{\cmark} \\ 
\bottomrule
\multicolumn{9}{l}{\footnotesize $^{\ast}$Verified by humans but static/outdated (no live web retrieval).}
\end{tabular}%
}
\end{table*}

\subsubsection{Dataset Statistics}
\label{subsubsec:stats}
\textbf{Component Distribution:} The dataset is stratified into three major subsets:
\begin{enumerate}
    \item \textbf{Fine-Grained Identification:} The largest component, comprising 48,580 images spanning 2,956 distinct species. This subset is organized into nine categories (e.g., \textit{Cereals}, \textit{Legumes}, \textit{Medicinal}) to ensure taxonomic breadth.
    \item \textbf{Disease Classification:} Consisting of 49,348 images, this subset targets 110 specific disease conditions across 33 major crops (averaging $\sim$1,701 images per crop). It covers critical staples such as \textit{Wheat}, \textit{Rice}, and \textit{Corn}, alongside cash crops like \textit{Coffee} and \textit{Sugarcane}.
    \item \textbf{Crop Counting \& Detection:} A specialized set of 23,497 images across 33 crops (averaging $\sim$712 images per crop), designed for spatial reasoning tasks.
\end{enumerate}

\begin{figure}[htbp]
    \centering
    \begin{subfigure}[b]{0.85\textwidth}
        \centering
        \includegraphics[width=\textwidth]{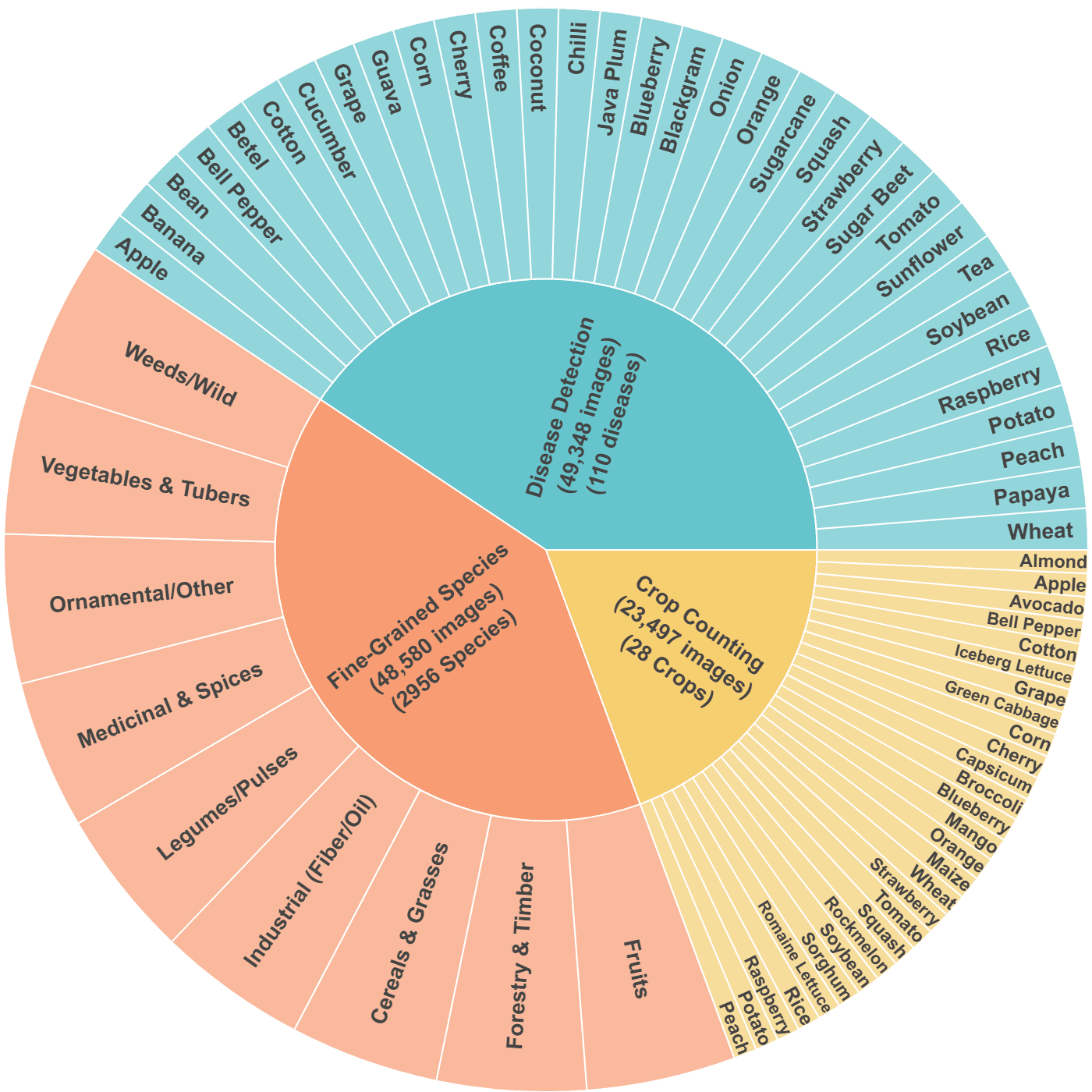}
        \caption{Hierarchical composition of AgriMM.}
        \label{fig:sunburst}
    \end{subfigure}
    
    \vspace{0.8em}
    
    \begin{subfigure}[b]{0.85\textwidth}
        \centering
        \includegraphics[width=\textwidth]{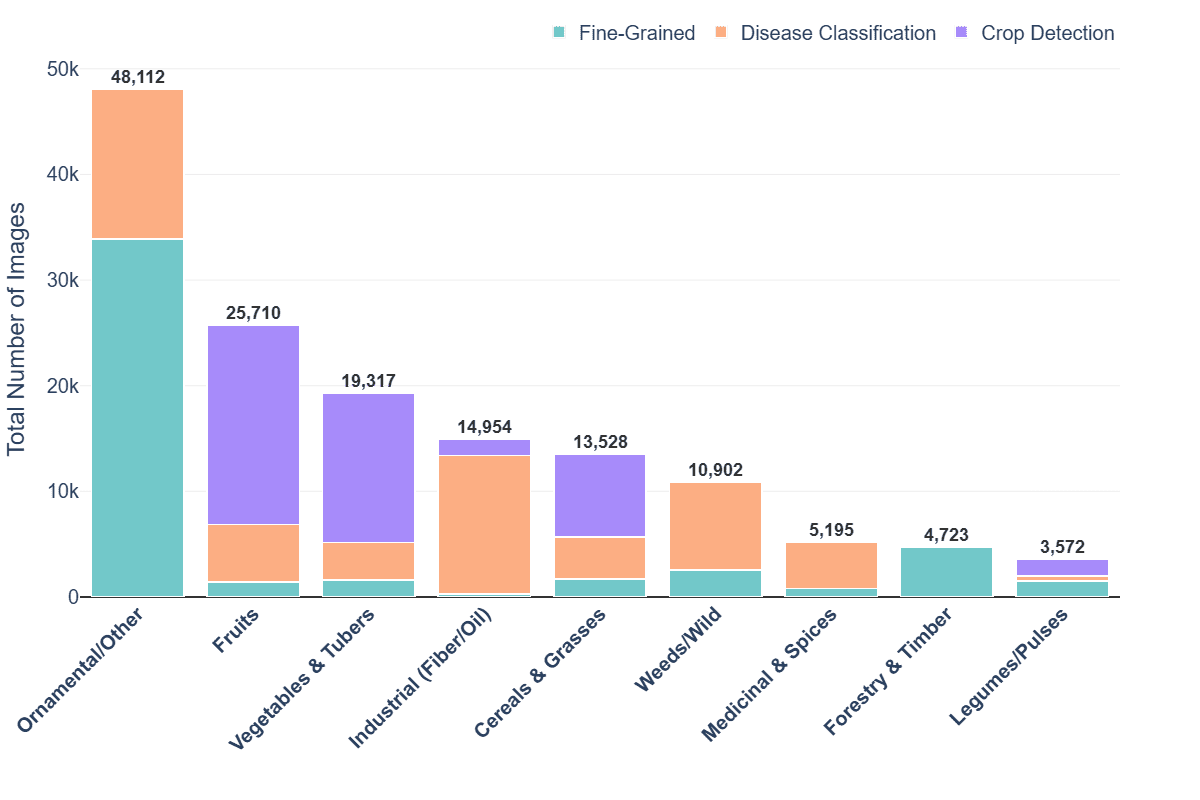}
        \caption{Quantitative distribution across categories.}
        \label{fig:bar_dist}
    \end{subfigure}
    \caption{AgriMM dataset statistics showing (a) taxonomic hierarchy and (b) class balance across functional categories.}
    \label{fig:dataset_stats}
\end{figure}

\textbf{Categorical Aggregation:} By integrating these three sources, AgriMM achieves a robust distribution across nine agricultural categories. The aggregated totals reveal a focus on high-economic value and high-biodiversity segments: Ornamental/Other (48,112 images), Fruits (25,710), Vegetables \& Tubers (19,317), and Industrial Crops (14,954). The remaining distribution ensures ecological completeness with Cereals \& Grasses (13,528), Weeds/Wild (10,902), Medicinal \& Spices (5,195), Forestry \& Timber (4,723), and Legumes/Pulses (3,572). This distribution (visualized in Figure~\ref{fig:dataset_stats}) prevents the common bias towards solely commercial crops by maintaining significant representation of wild and forestry species.

\subsubsection{Dataset Tasks}
\label{subsubsec:tasks}

AgriMM is designed to train models capable of addressing three critical agricultural AI capabilities, each reflecting real-world deployment requirements:

\noindent\textbf{1. Differential Pathology Diagnosis.}
Moving beyond binary healthy/sick classification, this task requires models to distinguish between visually similar conditions across 29 geographically diverse disease datasets. The collection spans multiple continents, from bean rust in Uganda to guava canker in Pakistan to vine viruses in Europe, ensuring models learn to generalize across environmental and cultivar variations. Critical diagnostic challenges include differentiating between overlapping symptomatology (e.g., \textit{Bacterial Spot} vs. \textit{Fungal Blight} in tomatoes), identifying co-occurring stresses (e.g., nutritional deficiency vs. pathogen infection in chilli), and classifying physical damage distinct from biological stress (e.g., 'crushed' vs. 'cracked' sugarcane stalks). This capability is essential for precision agriculture decision support systems where misdiagnosis can lead to incorrect treatment protocols.

\noindent\textbf{2. Species-Level Taxonomic Recognition.}
The fine-grained identification task challenges models to perform botanical discrimination using standardized scientific nomenclature rather than common names. Sourced from iNaturalist Agriculture (iNatAg), this subset includes observations from every continent, reflecting real-world environmental variability, from \textit{Abies balsamea} (balsam fir) in North American boreal forests to \textit{Acacia erioloba} (camelthorn) in African savannas. The task requires distinguishing between morphologically similar species within economically important genera (\textit{Acacia}, \textit{Eucalyptus}) and identifying invasive or noxious weeds in complex field backgrounds (e.g., \textit{Abutilon theophrasti} among cash crops). This capability supports biodiversity monitoring, invasive species detection, and regulatory compliance in agricultural trade.

\noindent\textbf{3. Quantitative Spatial Reasoning for Yield Estimation.}
Unlike classification tasks, counting and detection require models to perform spatial analysis under highly variable field conditions. The detection subset includes annotations for fruits and crops captured at different times of day (morning/night illumination), maturity stages (unripe/ripe/occluded strawberries), and density levels (sparse to heavily overlapping). Models must detect and count individual agricultural units, wheat heads in dense canopies, apples with varying occlusion, tomatoes at mixed ripeness, while reasoning about spatial relationships and yield potential. This capability directly supports automated harvest planning, quality grading systems, and economic forecasting for growers.

\subsubsection{Vision-to-Verified-Knowledge Synthesis}
\label{subsubsec:pipeline}

\begin{figure}[htbp]
    \centering
    \includegraphics[width=\linewidth]{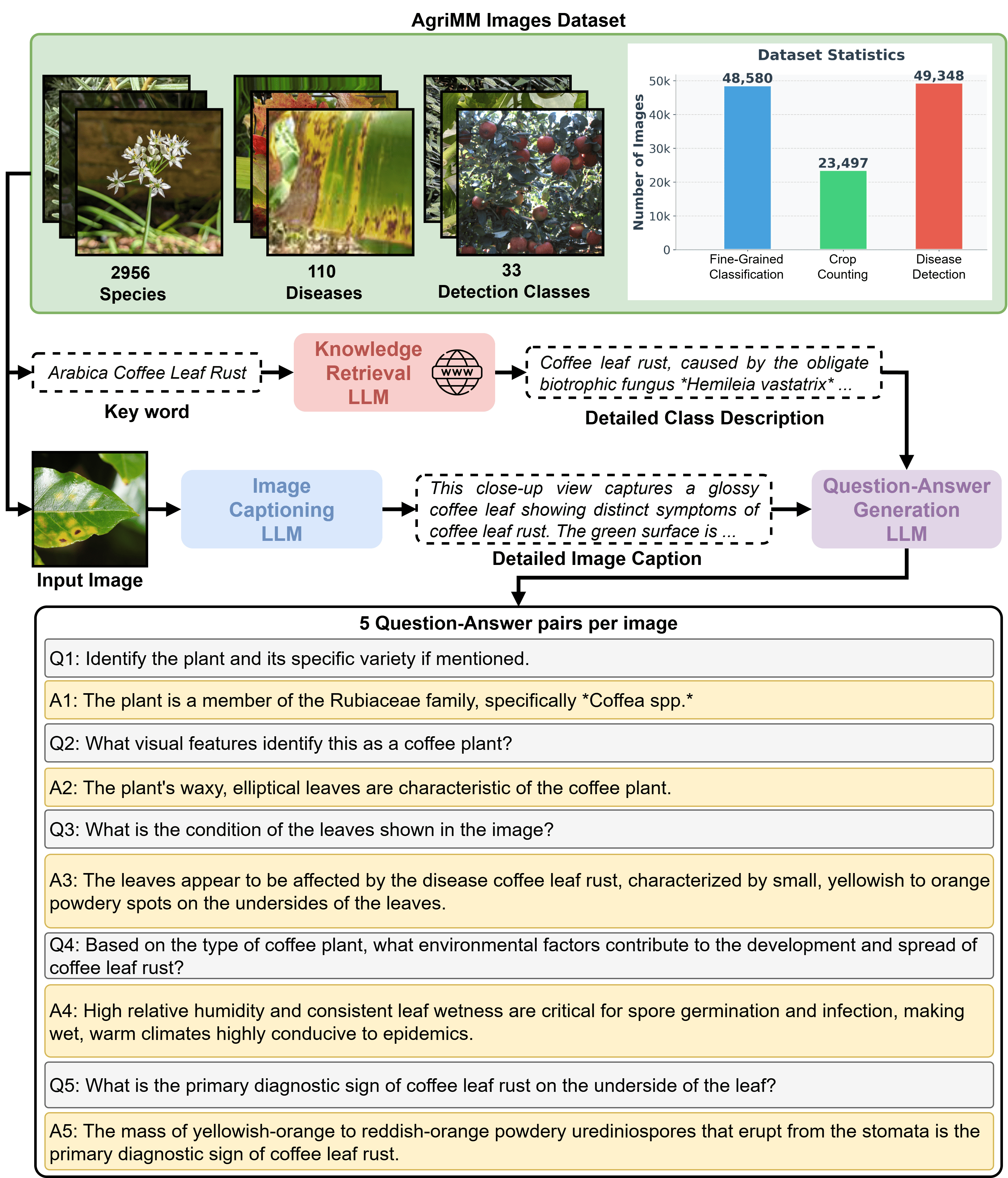}
    \caption{Overview of the Vision-to-Verified-Knowledge Synthesis Pipeline. Visual features are grounded in verified scientific literature through a three-stage generation and verification process.}
    \label{fig:pipeline}
\end{figure}

Generating high-quality instruction-following data at scale requires overcoming the biological hallucinations common in frozen language models. We introduce a three-stage \textbf{Vision-to-Verified-Knowledge (V2VK)} pipeline (Figure~\ref{fig:pipeline}) that grounds visual observations in scientifically verified literature (full prompts provided in ~\ref{app:prompts}).

\textbf{Stage I: Visual Grounding via Image Captioning.}
We employ Gemma 3 (12B)~ \cite{gemma2024} to generate structured natural language descriptions. To prevent label drift, we condition the generation on the ground-truth labels from AgriMM (e.g., injecting ``\textit{Solanum lycopersicum}''). The prompt enforces the extraction of agricultural metadata, including growth stage, planting density, and environmental context.

\textbf{Stage II: Knowledge Retrieval and Verification.}
Visual captions lack deep botanical knowledge. We address this using Gemini 3 Pro~ \cite{gemini2024} with a Retrieval-Augmented Generation (RAG) approach. The model retrieves contemporary scientific descriptions, management protocols, and phenological data from trusted databases.

\textbf{Stage III: Instruction Synthesis.}
The final stage synthesizes the visual captions (Stage I) and verified knowledge (Stage II) into 607,125 diverse QA pairs using LLaMA-3.1-8B-Instruct \cite{llama3-2024}. We utilize a constrained prompting strategy that mandates five distinct question types per image: Identification, Visual Reasoning, Health Condition, Cultivation Knowledge, and Quantification. For images belonging to the Crop Counting \& Detection subset, the ground-truth crop count is derived by tallying the number of bounding box annotations per image and injected directly into the generation prompt. This ensures that Quantification QA pairs reflect verifiable, annotation-grounded values rather than model estimates, preventing the hallucination of incorrect counts.

\subsubsection{Pipeline Verification and Quality Assurance}
\label{subsubsec:verification}

To guarantee the biological validity and semantic accuracy of the generated instructions, we implemented a rigorous human-in-the-loop verification protocol. While the pipeline automates the synthesis of scale, manual oversight is critical to prevent the propagation of hallucinations common in Large Language Models.

\textbf{Manual Knowledge verification:} We conducted a comprehensive review of the scientific knowledge retrieved in Stage II. Expert annotators manually fact-checked the detailed class descriptions against established agricultural taxonomies and phytopathology literature. This process ensured that the retrieved biological traits, disease symptoms, and phenological data accurately corresponded to the visual evidence (Stage I).

\subsection{AgriChat Architecture}
\label{sec:architecture}

\begin{figure}[htbp]
    \centering
    \includegraphics[width=\textwidth]{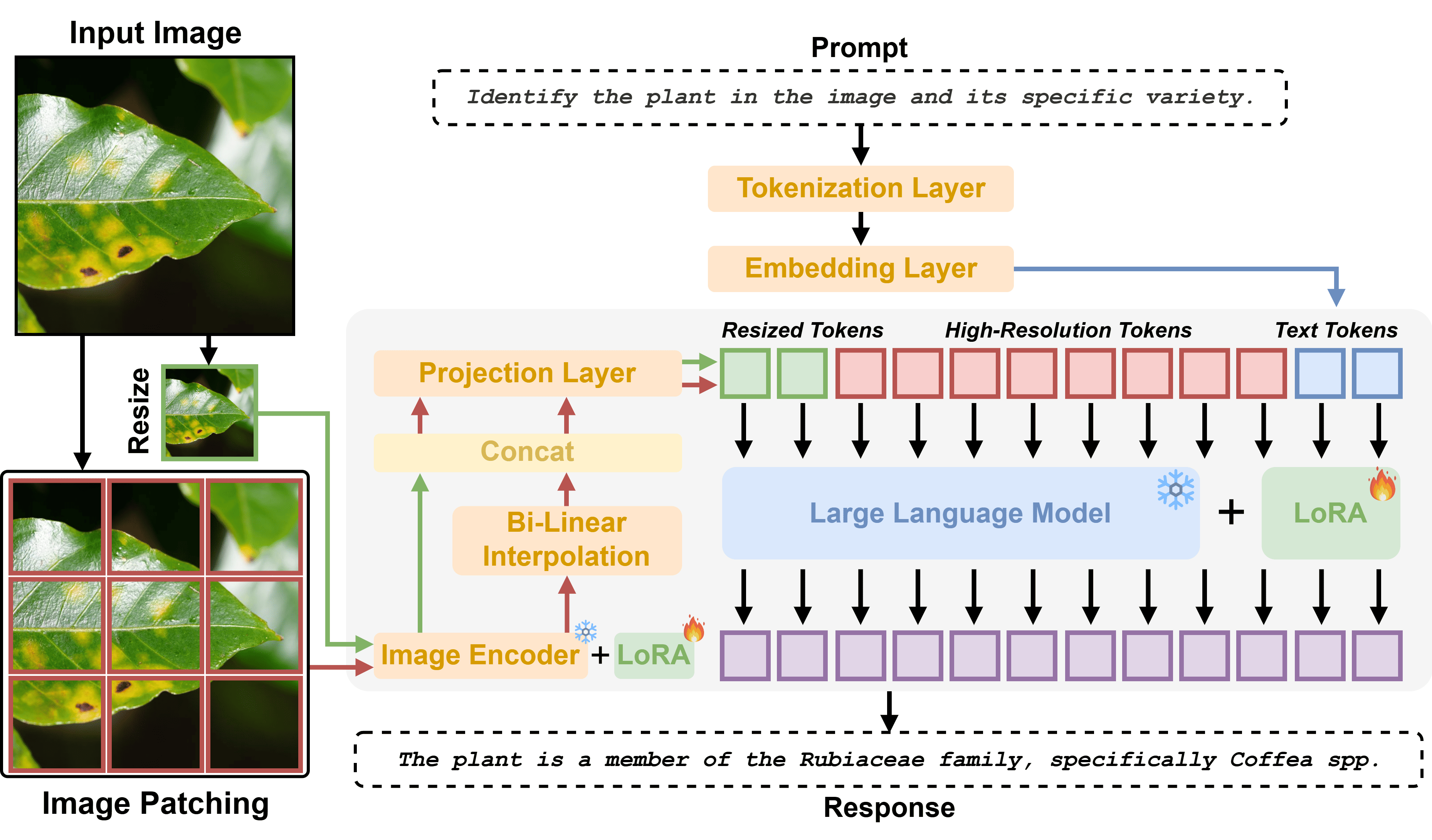} 
    \caption{\textbf{Overview of the AgriChat Architecture.} The model utilizes an adaptive resolution strategy to process high-quality agricultural imagery. Input images are split into local patches and a resized global thumbnail, encoded by SigLIP equipped with LoRA adapters, and aligned via a projection layer. The visual tokens are concatenated with text instructions and processed by LLM backbone equipped with LoRA adapters (SwiGLU + Multi-Head Self-Attention) to generate verifiable diagnostic responses.}
    \label{fig:architecture}
\end{figure}

As illustrated in Figure~\ref{fig:architecture}, AgriChat is a domain-specialized Multimodal Large Language Model (MLLM) tailored for high-precision agricultural diagnostics. The architecture is composed of four primary modules: a high-resolution vision encoder, a cross-modal projection network, a Large Language Model (LLM) decoder, and a parameter-efficient adaptation mechanism. In contrast to broad-scope MLLMs that prioritize conversational versatility, AgriChat focuses on fine-grained visual discrimination, a prerequisite for detecting minute pest damage, early-onset pathologies, and subtle phenotypic traits.

\subsubsection{Vision Encoder: SigLIP-SO400M}

For visual feature extraction, we utilize SigLIP-SO400M (Sigmoid Loss for Language-Image Pre-training)~ \cite{siglip2023}. This contrastive model was chosen over standard CLIP variants for its superior ability to preserve spatial details essential for agricultural tasks.

Formally, let $\mathcal{T} \in \mathbb{R}^{H_v \times W_v \times 3}$ be an input image tile, where the native resolution is fixed at $H_v = W_v = 384$. The encoder, denoted as $\mathcal{E}_{\omega + \Delta\omega}$, retains its pre-trained base parameters $\omega$ in a frozen state while introducing a set of trainable LoRA parameters $\Delta\omega$ to adapt the encoder to agricultural visual features. It processes the tile to produce a spatial feature map $\mathbf{F}_{sp}$, defined as:
\begin{equation}
\mathbf{F}_{sp} = \mathcal{E}_{\omega + \Delta\omega}(\mathcal{T}) \in \mathbb{R}^{h \times w \times d_v}
\end{equation}
where $h=w=27$ represents the spatial grid size, and $d_v = 1152$ is the visual hidden dimension. Flattening this map yields a sequence of $N_v = h \times w = 729$ tokens per tile.

\subsubsection{High-Resolution Visual Encoding}
\label{subsubsec:anyres}

To mitigate the resolution loss common in standard resizing, we adopt the adaptive resolution strategy~ \cite{llava-onevision2024}. This allows the model to process high-resolution field photography by decomposing images into native-resolution patches.

\noindent\textbf{Stage 1: Adaptive Grid Generation.} Given a high-resolution input image $\mathcal{I} \in \mathbb{R}^{H \times W \times 3}$, we calculate a grid configuration $(n_h, n_w)$ that best approximates the original aspect ratio. This grid is constrained by a pre-defined token budget $N_{\max}$. In our implementation, we set $N_{\max} = 8748$, allowing the processing of images up to $1344 \times 1344$ resolution.

\noindent\textbf{Stage 2: Dual-Path Feature Extraction.} The encoding process operates on two parallel tracks:
\begin{itemize}
    \item \textbf{Local Path:} The image is divided into $n_h \times n_w$ non-overlapping patches $\{\mathcal{T}_{i,j}\}$. Each patch is encoded individually and flattened to extract local details: $\mathbf{H}_{i,j} = \text{Flatten}(\mathcal{E}_{\omega + \Delta\omega}(\mathcal{T}_{i,j})) \in \mathbb{R}^{N_v \times d_v}$.
    \item \textbf{Global Path:} A downsampled thumbnail $\mathcal{T}_{\text{global}}$ is encoded to capture holistic context: $\mathbf{H}_{\text{global}} = \text{Flatten}(\mathcal{E}_{\omega + \Delta\omega}(\mathcal{T}_{\text{global}})) \in \mathbb{R}^{N_v \times d_v}$.
\end{itemize}

\noindent\textbf{Stage 3: Adaptive Pooling.} 
Let $L$ denote the total number of visual tokens generated by the dual paths, calculated as $L = (n_h \cdot n_w + 1) \cdot N_v$. To handle variable resolutions efficiently, if $L$ exceeds the sequence limit $N_{\max}$, we apply bilinear interpolation, denoted as $\text{Interp}(\cdot)$, in the feature space. Let $\mathbf{F}_{i,j} \in \mathbb{R}^{h \times w \times d_v}$ be the unflattened feature map of a specific tile. The final local features $\tilde{\mathbf{H}}_{i,j}$ are computed as:
\begin{equation}
\tilde{\mathbf{H}}_{i,j} = \begin{cases} 
\text{Flatten}\left( \text{Interp}(\mathbf{F}_{i,j}, \text{size}=(h', w')) \right) & \text{if } L > N_{\max} \\
\mathbf{H}_{i,j} & \text{otherwise}
\end{cases}
\end{equation}
where $(h', w')$ represents the reduced spatial dimensions calculated to satisfy the condition that the new total token count $\tilde{L} \leq N_{\max}$.

\subsubsection{Cross-Modal Projection Network}

To bridge the modality gap between the visual encoder ($d_v = 1152$) and the language model ($d_{llm} = 3584$), we employ a Multi-Layer Perceptron (MLP) projector $\mathcal{P}_{\psi}$. We use the bracket notation $[\cdot, \cdot]$ to denote tensor concatenation along the sequence dimension. 

The complete sequence of visual features is formed by concatenating the global thumbnail features with the processed local patch features $\tilde{\mathbf{H}}_{i,j}$:
\begin{equation}
\mathbf{H}_{\text{visual}} = [\mathbf{H}_{\text{global}}, \tilde{\mathbf{H}}_{1,1}, \dots, \tilde{\mathbf{H}}_{n_h, n_w}] \in \mathbb{R}^{\tilde{L} \times d_v}
\end{equation}
Let $\psi = \{\mathbf{W}_1, \mathbf{b}_1, \mathbf{W}_2, \mathbf{b}_2\}$ be the trainable projection weights, and $\text{GELU}(\cdot)$ be the Gaussian Error Linear Unit activation function. The projected visual tokens $\mathbf{Z}_v$ are obtained via:
\begin{equation}
\mathbf{Z}_v = \mathcal{P}_{\psi}(\mathbf{H}_{\text{visual}}) = \text{GELU}(\mathbf{H}_{\text{visual}} \mathbf{W}_1 + \mathbf{b}_1)\mathbf{W}_2 + \mathbf{b}_2
\end{equation}
This results in $\mathbf{Z}_v \in \mathbb{R}^{\tilde{L} \times d_{llm}}$, aligning the visual representations with the LLM's embedding space.

\subsubsection{Language Model Decoder}

The core reasoning engine is Qwen-2-7B~ \cite{qwen2024}, a transformer-based decoder $\mathcal{D}_{\phi + \Delta\phi}$ whose pre-trained weights $\phi$ are kept frozen while trainable LoRA parameters $\Delta\phi$ are injected to enable domain adaptation.

\textbf{Multimodal Fusion:} Fusion is achieved through early embedding integration \cite{llava-onevision2024}. Let $\mathcal{E}_{\text{text}}(\cdot)$ denote the word embedding layer of the LLM which maps discrete tokens to $\mathbb{R}^{d_{llm}}$. For a tokenized instruction sequence $\mathbf{X}_{\text{text}} = \{x_1, \dots, x_n\}$ containing a special placeholder token $x_k = \texttt{<image>}$, the input embeddings $\mathbf{S}$ are created by replacing the placeholder with the visual sequence:
\begin{equation}
\mathbf{S} = [\mathcal{E}_{\text{text}}(x_1), \dots, \mathcal{E}_{\text{text}}(x_{k-1}), \mathbf{Z}_v, \mathcal{E}_{\text{text}}(x_{k+1}), \dots, \mathcal{E}_{\text{text}}(x_n)]
\end{equation}
This unified sequence $\mathbf{S}$ serves as the input to the decoder.

\subsection{Training Strategy}

\subsubsection{Parameter-Efficient Domain Adaptation (LoRA)}
To adapt the model to the agricultural domain, we utilize Low-Rank Adaptation (LoRA)~ \cite{hu2022lora} across both the vision encoder and the language model decoder. This strategy keeps all pre-trained base parameters frozen while injecting lightweight, trainable low-rank adapters, enabling efficient domain specialization with minimal computational overhead.

\paragraph{Vision Encoder Adaptation.}
The pre-trained SigLIP-SO400M \cite{siglip2023} parameters $\omega$ are frozen, and trainable LoRA adapters $\Delta\omega$ are injected into the linear layers of the vision encoder. For a frozen linear layer defined by weights $\mathbf{W}_0^{(\text{vis})} \in \mathbb{R}^{D_{in}^{(\text{vis})} \times D_{out}^{(\text{vis})}}$, we define adapter matrices $\mathbf{A}^{(\text{vis})} \in \mathbb{R}^{D_{in}^{(\text{vis})} \times r_v}$ and $\mathbf{B}^{(\text{vis})} \in \mathbb{R}^{r_v \times D_{out}^{(\text{vis})}}$, where $r_v = 32$ is the low-rank dimension for the vision encoder. The forward pass for an input $\mathbf{x} \in \mathbb{R}^{1 \times D_{in}^{(\text{vis})}}$ is modified as:
\begin{equation}
\mathbf{h}^{(\text{vis})} = \mathbf{x} \mathbf{W}_0^{(\text{vis})} + \mathbf{x} \mathbf{A}^{(\text{vis})} \mathbf{B}^{(\text{vis})} \cdot \frac{\alpha_v}{r_v}
\end{equation}
where $\alpha_v$ is the scaling factor for the vision encoder adapters.

\paragraph{Language Model Adaptation.}
We inject trainable Low-Rank Adapters into the linear layers of the decoder (specifically the Attention and SwiGLU projection layers). The base LLM parameters $\phi$ are kept frozen in their native precision. Let $D = 3584$ be the hidden dimension of the Qwen-2-7B backbone. For a frozen linear layer defined by weights $\mathbf{W}_0 \in \mathbb{R}^{D_{in} \times D_{out}}$, we define adapter matrices $\mathbf{A} \in \mathbb{R}^{D_{in} \times r}$ and $\mathbf{B} \in \mathbb{R}^{r \times D_{out}}$, where $r=128$ is the low-rank dimension. The forward pass for an input $\mathbf{x} \in \mathbb{R}^{1 \times D_{in}}$ is modified as follows:
\begin{equation}
\mathbf{h} = \mathbf{x} \mathbf{W}_0 + \mathbf{x} \mathbf{A} \mathbf{B} \cdot \frac{\alpha}{r}
\end{equation}
where $\alpha=256$ is the scaling factor. For both components, $\mathbf{A}$ matrices are initialized from a Gaussian distribution, while $\mathbf{B}$ matrices are initialized to zero to ensure training starts from the identity function.

\subsubsection{Optimization Objective}
The model is optimized using the standard autoregressive cross-entropy loss. Let $\Theta = \psi \cup \{\mathbf{A}_l^{(\text{vis})}, \mathbf{B}_l^{(\text{vis})}\}_{\forall l} \cup \{\mathbf{A}_l, \mathbf{B}_l\}_{\forall l}$ represent the union of all trainable parameters (projector weights and LoRA adapters across both the vision encoder and language model decoder). 

Given a ground-truth response sequence $\mathbf{Y}$ of length $T$, we apply a mask $M_t \in \{0, 1\}$ to exclude instruction tokens from the loss calculation. The optimization objective is defined as:
\begin{equation}
\mathcal{L}(\Theta) = - \sum_{t=1}^{T} M_t \log P(y_t \mid \mathbf{Y}_{<t}, \mathbf{S}; \Theta)
\end{equation}
This ensures the gradients are backpropagated only through $\Theta$ based on the likelihood of the correct agricultural diagnostic tokens $y_t$.

\section{Experimental Evaluations}
\label{sec:results}

\subsection{Training and Implementation Details}
\label{subsec:training_details}
AgriChat was fine-tuned exclusively on the AgriMM dataset using a consumer-grade NVIDIA RTX 3090 GPU (24GB VRAM). The AgriMM dataset was partitioned using an 80:20 train/test split, with no overlap between the training and evaluation subsets. Fine-tuning was performed for 1 epoch with a per-device batch size of 1 and gradient accumulation steps of 16, yielding an effective batch size of 16. A learning rate of $2 \times 10^{-4}$ was used with bfloat16 mixed precision. LoRA adapters were applied to both the Qwen2 LLM backbone (rank $r=128$, $\alpha=256$) and the SigLIP vision encoder (rank $r=32$, $\alpha=64$), targeting the attention projections (\texttt{q\_proj}, \texttt{k\_proj}, \texttt{v\_proj}) in both components, the LLM-specific layers (\texttt{o\_proj}, \texttt{gate\_proj}, \texttt{up\_proj}, \texttt{down\_proj}), and the vision encoder-specific layers (\texttt{out\_proj}, \texttt{fc1}, \texttt{fc2}). Inference benchmarks were conducted using 4-bit quantization, a batch size of 1, and a standardized test image (1,328 KB) with the prompt \textit{``What is the name of the plant in the image?''} repeated across 100 inference cycles. Given that comparable domain-specific models such as Agri-LLaVA and AgroGPT are not publicly released as of this writing, AgriChat (7B) was evaluated against three state-of-the-art generalist multimodal models: LLaVA-OneVision (7B) \cite{llava-onevision2024}, Llama-3.2 (Vision-11B) \cite{llama3-2024}, and Qwen-2.5 (VL-7B) \cite{qwen2024}. These baselines were selected as representative of the current frontier in open-weight vision-language models across a range of parameter scales.

\subsection{Evaluation Metrics}
\label{subsec:evaluation_metrics}

Evaluating generative conversational agents requires a multi-faceted framework, as no single metric captures all dimensions of response quality~ \cite{reiter2018nlg, liu2023geval}. We therefore combine lexical overlap metrics, semantic similarity measures, and a multimodal alignment score.

\subsubsection{Lexical and N-Gram Metrics}
\label{subsubsec:lexical_metrics}

Lexical metrics measure surface-level textual overlap and provide a standard baseline for text generation evaluation:

\begin{itemize}
    \item \textbf{BLEU-4:} Measures $n$-gram precision (unigrams through 4-grams) against the reference, with a brevity penalty~ \cite{papineni2002bleu}.

    \item \textbf{ROUGE-2:} Measures bigram overlap between system output and reference, capturing basic phrasal coherence~ \cite{lin2004rouge}.
    
    \item \textbf{METEOR:} Extends exact matching with stemming and synonymy via WordNet, and penalizes word order violations~ \cite{banerjee2005meteor}.
\end{itemize}

These metrics are nonetheless limited by their semantic blindness: a model generating a paraphrase of the correct answer is penalized despite factual equivalence~ \cite{nainia2025broken}. This motivates the semantic and judge-based metrics below.

\subsubsection{Semantic and Embedding-Based Metrics}
\label{subsubsec:semantic_metrics}

\begin{itemize}
    \item \textbf{BERTScore (F1):} Computes token-level cosine similarity between candidate and reference using contextual BERT embeddings~ \cite{zhang2019bertscore}.

    \item \textbf{Long-CLIP Cosine Similarity:} Evaluates semantic alignment between generated text and reference text within CLIP's shared embedding space~ \cite{radford2021clip}, providing a text-to-text similarity measure that captures conceptual and semantic relationships beyond simple lexical overlap.

    \item \textbf{T5 Cosine Similarity (T5 Cos):} Measures semantic similarity using the T5 encoder's latent representations~ \cite{raffel2020t5}.
    
    \item \textbf{SBERT Similarity:} Computes cosine similarity between sentence-level dense embeddings from a Siamese BERT architecture~ \cite{reimers2019sbert}.
\end{itemize}

\subsubsection{LLM-as-a-Judge}
\label{subsubsec:llm_judge}

Our analysis reveals a critical limitation of the lexical and embedding-based metrics described above, particularly salient for conversational diagnostic agents: a systematic bias toward brevity, which we term the \textit{verbosity penalty}. This phenomenon results in misleadingly low scores for models providing deeper reasoning, even when diagnostic accuracy remains correct. This issue arises from a mismatch between reference formatting and assistant-optimized generation. Ground truth annotations in standard diagnostic datasets typically provide minimal, binary responses (e.g., ``No''), whereas models trained for conversational assistance generate contextual explanations.

\begin{quote}
\textbf{Example of Metric Misalignment:} \\
\textit{Prompt:} ``Can you detect signs indicating absence of pathogens?'' \\
\textit{Ground Truth:} ``No'' \\
\textit{Concise Prediction:} ``No'' $\rightarrow$ \textbf{BERTScore: 1.0} \\
\textit{Verbose Prediction:} ``No, the image does not provide information about the presence or absence of pathogens.'' $\rightarrow$ \textbf{BERTScore: $\approx$ 0.50}
\end{quote}

Both predictions are factually correct and clinically equivalent, yet surface-level metrics heavily penalize the more informative response. This is not an isolated observation: \cite{liu2023geval} have formally demonstrated that conventional reference-based metrics such as BLEU and ROUGE exhibit relatively low correlation with human judgments, particularly on tasks requiring contextual, open-ended generation—precisely the setting of our diagnostic assistant. This limitation is further corroborated by \cite{zheng2023judging}, who identify \textit{verbosity bias} as a documented failure mode in automated evaluation, wherein longer but substantively equivalent responses receive systematically lower scores despite no loss in factual accuracy. More broadly, \cite{gu2024llmsasjudges} highlight in their comprehensive survey that traditional metrics fail to capture key aspects such as logical coherence and semantic correctness in generative tasks, motivating a paradigm shift toward LLM-based evaluation. These converging findings motivate the adoption of the \textit{LLM-as-a-Judge} paradigm~ \cite{llm_as_a_Judge}, which enables evaluation based on semantic intent rather than surface form and correlates more strongly with human judgment than n-gram matching. We employ Qwen3-30B-A3B-Instruct as our evaluator. Our selection is grounded in the ``Judge's Verdict Benchmark'' recently introduced by NVIDIA \cite{judges_verdict}. In their extensive analysis of 54 LLM judges, Qwen3-30B was identified as a Tier 1 evaluator, demonstrating exceptional alignment with human annotators. Specifically, it achieved a Z-score of $|z| = 0.04$ in the ``Turing Test for Judges,'' classifying it as a \textit{Human-Like Judge} capable of preserving the nuances of expert evaluation, outperforming larger proprietary models in agreement stability. Following the prompt design strategies outlined in \cite{llm_as_a_Judge} regarding \textit{Criteria Decomposition}, our evaluation prompt separates assessment into four distinct axes: Correctness, Completeness, Clarity, and Conciseness (full prompt can be found in \ref{app:eval_prompts}). We utilize a 4-point Likert scale (1--4) rather than the traditional 5-point scale. This design choice draws upon the methodology in \cite{judges_verdict}, which emphasizes the utility of discrete, normalized scoring to improve reliability. A 4-point scale forces the evaluator to make a definitive decision (positive vs. negative) by removing the neutral middle option, thereby reducing the central tendency bias often observed in automated judges. By leveraging a model proven to exhibit human-like agreement patterns \cite{judges_verdict} and a constrained scoring rubric, we ensure that higher scores reflect genuine diagnostic utility rather than stochastic generative fluency.

\subsection{Evaluation Datasets}
\label{subsec:datasets}
Our experimental framework evaluates performance across a diverse set of agricultural benchmarks, ranging from controlled laboratory settings to complex field environments.

\begin{enumerate}
    \item \textbf{AgriMM (Ours):} The primary training and in-domain evaluation set. It consolidates 63 distinct data sources into 121,425 images and 607,125 QA pairs, verified via a Vision-to-Verified-Knowledge pipeline. It covers fine-grained species identification, disease diagnosis, and crop counting.
    
    \item \textbf{PlantVillage \cite{hughes2015open}:} A foundational dataset comprising 54,306 images of 38 disease classes. We use this as a zero-shot transfer benchmark to evaluate performance on laboratory-controlled imagery.
    
    \item \textbf{CDDM (Crop Disease Diagnosis Multimodal) \cite{cddm}:} A large-scale dataset focusing on 16 major crop categories. We utilize this to test generalization on web-crawled field images.
    
    \item \textbf{AGMMU \cite{AGMMU}:} A benchmark derived from 58,571 expert-farmer dialogues. We utilize both the Multiple Choice Question (MCQ) and Open-Ended subsets to evaluate conversational reasoning and management advice capabilities.
\end{enumerate}

\subsection{Qualitative Analysis}
\label{subsec:qualitative}

To complement the quantitative evaluation, we present representative examples that illustrate the strengths and limitations of AgriChat relative to generalist baselines. These cases were selected to demonstrate key findings across a variety of agricultural vision-language tasks, including counting, species identification, open-ended decision-making, multiple-choice reasoning, and disease diagnosis. It is important to note that AgriChat was fine-tuned exclusively on the AgriMM dataset; all remaining examples reflect zero-shot generalization to unseen benchmarks.

\subsubsection{Example 1: Precise Agricultural Counting (AgriMM)}

\begin{figure}[htbp]
    \centering
    \includegraphics[width=\linewidth]{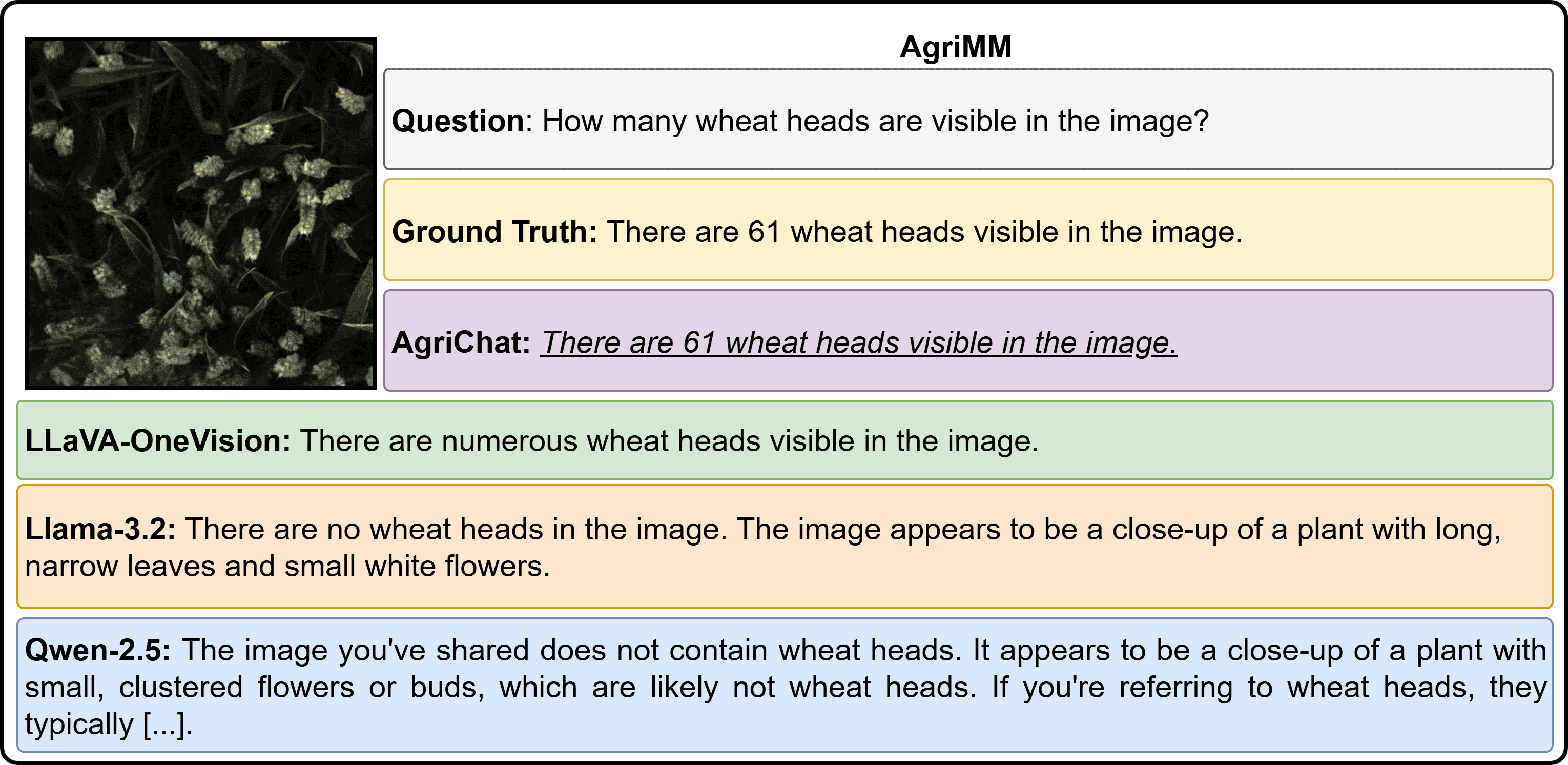}
    \caption{Qualitative comparison on an AgriMM wheat-head counting task.}
    \label{fig:qual1}
\end{figure}

As shown in Figure~\ref{fig:qual1}, AgriChat is the only model to produce the exact numerical answer, correctly identifying 61 wheat heads in the image. This reflects the direct benefit of fine-tuning on AgriMM, which exposes the model to precise, count-oriented agricultural queries. The generalist baselines fail substantially: LLaVA-OneVision resorts to a vague quantifier ("numerous"), while both Llama-3.2 and Qwen-2.5 incorrectly assert that no wheat heads are present at all, misidentifying the crop entirely. This stark contrast highlights how domain-specific fine-tuning equips AgriChat with the visual grounding necessary for fine-grained agricultural counting, a task that generalist models, trained without such priors, are ill-equipped to handle.

\subsubsection{Example 2: Zero-Shot Species Identification (PlantVillageVQA)}

\begin{figure}[htbp]
    \centering
    \includegraphics[width=\linewidth]{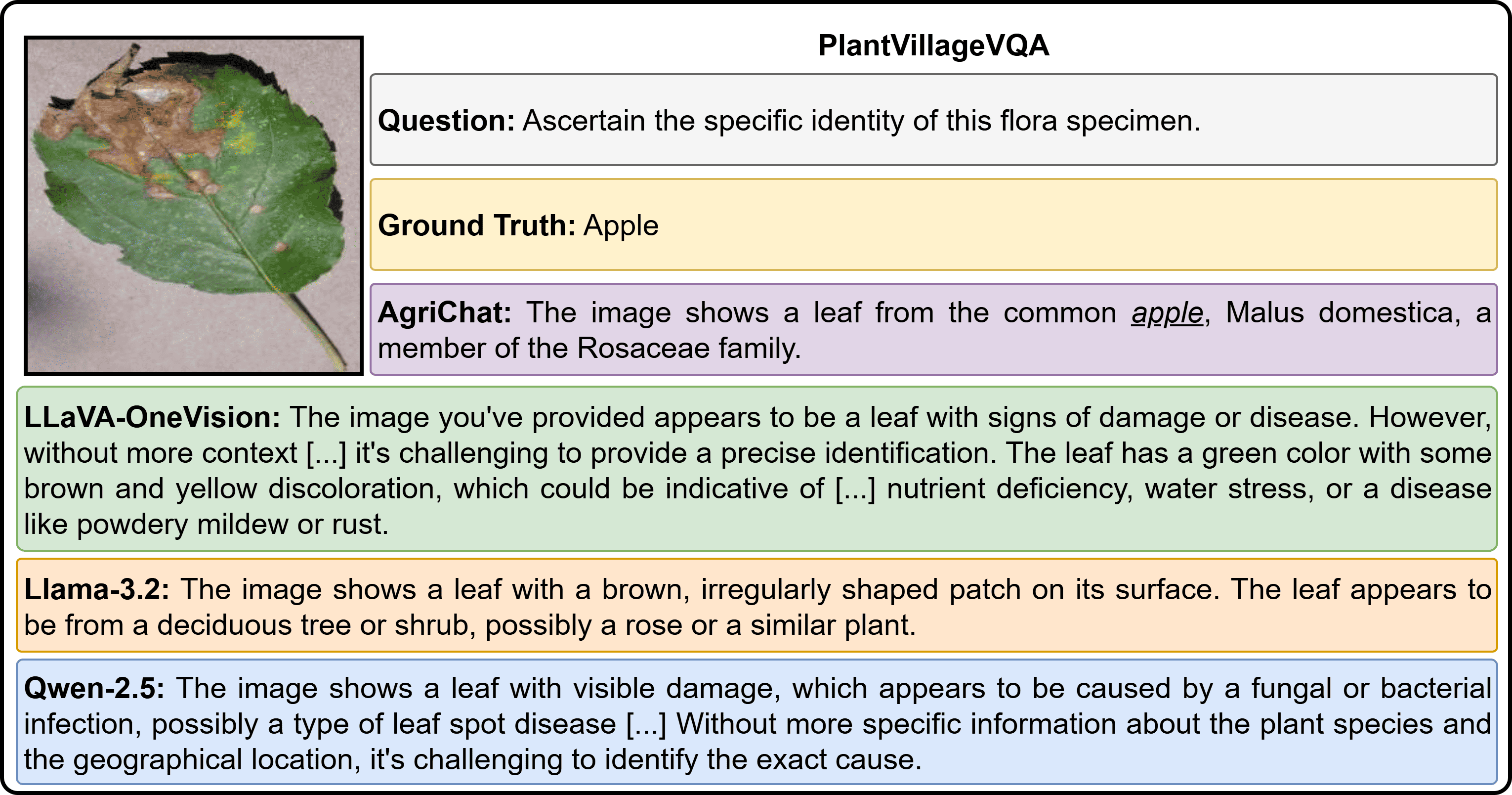}
    \caption{Qualitative comparison on a PlantVillageVQA plant identification task.}
    \label{fig:qual2}
\end{figure}

Figure~\ref{fig:qual2} illustrates a zero-shot species identification task in which AgriChat correctly identifies the specimen as \textit{Malus domestica} and situates it within the appropriate taxonomic family. The generalist baselines, by contrast, uniformly fixate on the visible pathology and treat the query as a disease diagnosis task rather than a species identification task. This reflects a fundamental misalignment between generalist model behavior and agricultural workflows, where correctly identifying ``what plant'' must logically precede determining ``what ails it.'' AgriChat's agricultural fine-tuning appears to instil the correct task prioritization even in zero-shot settings.

\subsubsection{Example 3: Open-Ended Management Reasoning (AGMMU)}

\begin{figure}[htbp]
    \centering
    \includegraphics[width=\linewidth]{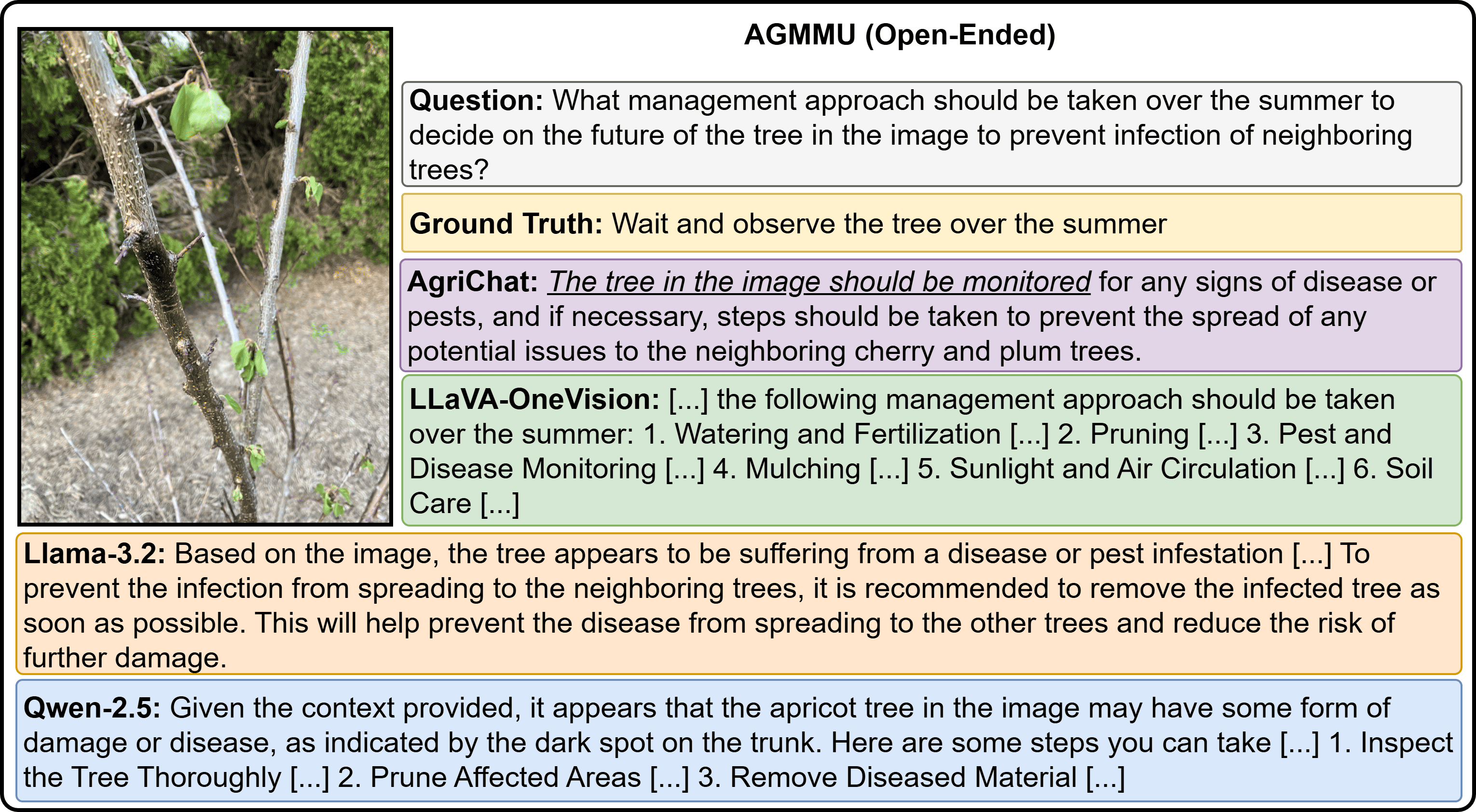}
    \caption{Qualitative comparison on an AGMMU open-ended tree management task.}
    \label{fig:qual3}
\end{figure}

The open-ended management scenario presented in Figure~\ref{fig:qual3} tests the models' ability to provide contextually appropriate, cautious advice under uncertainty. AgriChat's response aligns closely with the ground truth by recommending monitoring and watchful observation rather than immediate intervention, appropriately acknowledging the risk to neighboring trees. The generalist baselines lean in opposite directions: LLaVA-OneVision produces a generic horticultural checklist that ignores the specific decision context, while Llama-3.2 recommends immediate removal of the tree, a drastic and potentially unnecessary intervention. Qwen-2.5 similarly defaults to aggressive pruning and remediation steps. AgriChat's measured, context-sensitive response suggests that domain fine-tuning fosters a more calibrated reasoning style aligned with real-world agricultural decision-making.

\subsubsection{Example 4: Multiple-Choice Visual Reasoning (AGMMU)}

\begin{figure}[htbp]
    \centering
    \includegraphics[width=\linewidth]{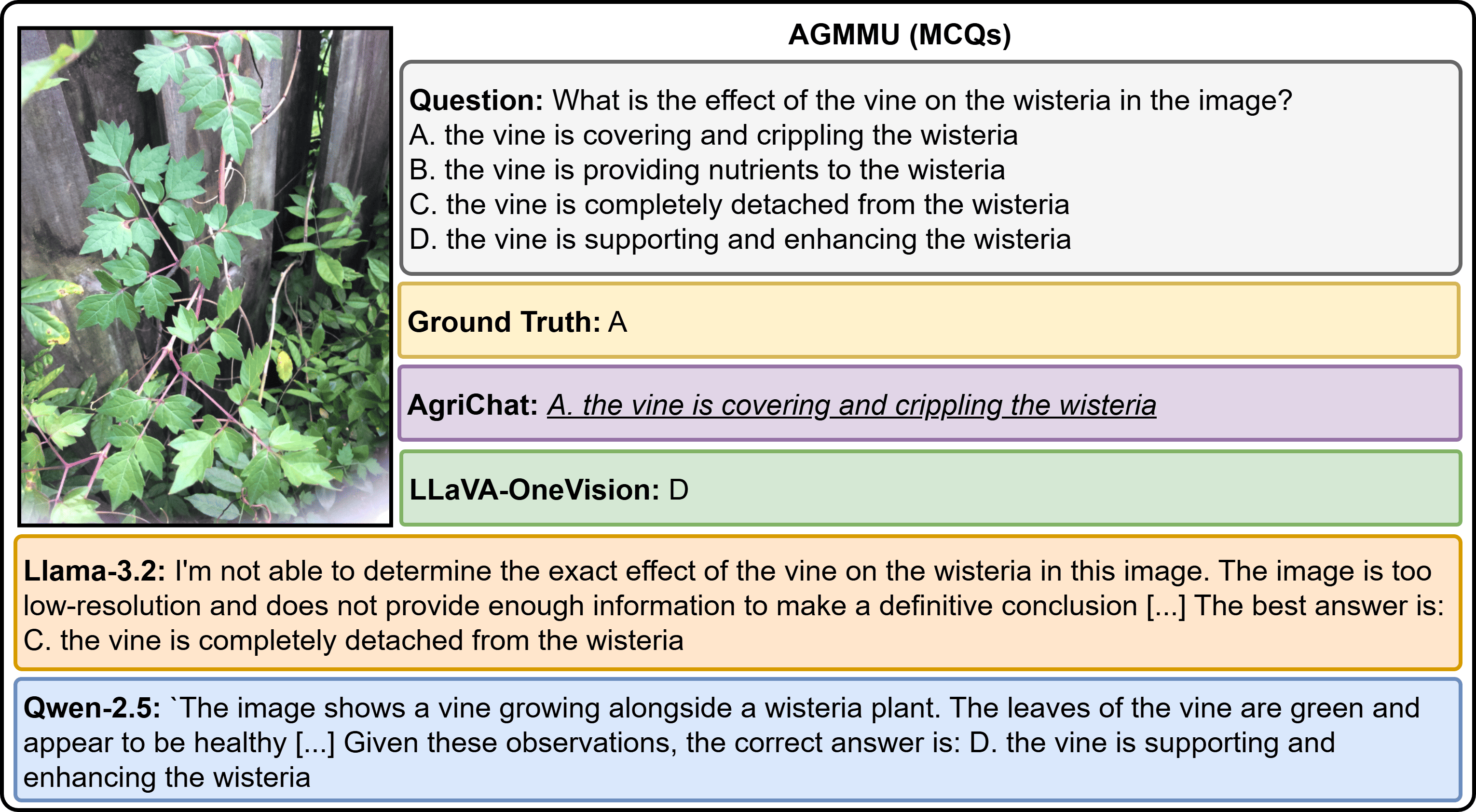}
    \caption{Qualitative comparison on an AGMMU multiple-choice visual reasoning task.}
    \label{fig:qual4}
\end{figure}

In the zero-shot multiple-choice task depicted in Figure~\ref{fig:qual4}, AgriChat selects the correct answer (A) while all three generalist baselines fail. LLaVA-OneVision selects option D without justification, Llama-3.2 cites low image resolution as an excuse before incorrectly choosing C, and Qwen-2.5 constructs a plausible-sounding but factually incorrect rationale in favour of D. AgriChat, despite not being trained on AGMMU, correctly interprets the visual relationship between the vine and the wisteria as detrimental, demonstrating that its agricultural priors transfer to structured reasoning tasks. The pattern of confident but incorrect responses from the baselines underscores the risk of deploying generalist models in agricultural advisory contexts without domain adaptation.

\subsubsection{Example 5: Disease Diagnosis (CDDM)}

\begin{figure}[htbp]
    \centering
    \includegraphics[width=\linewidth]{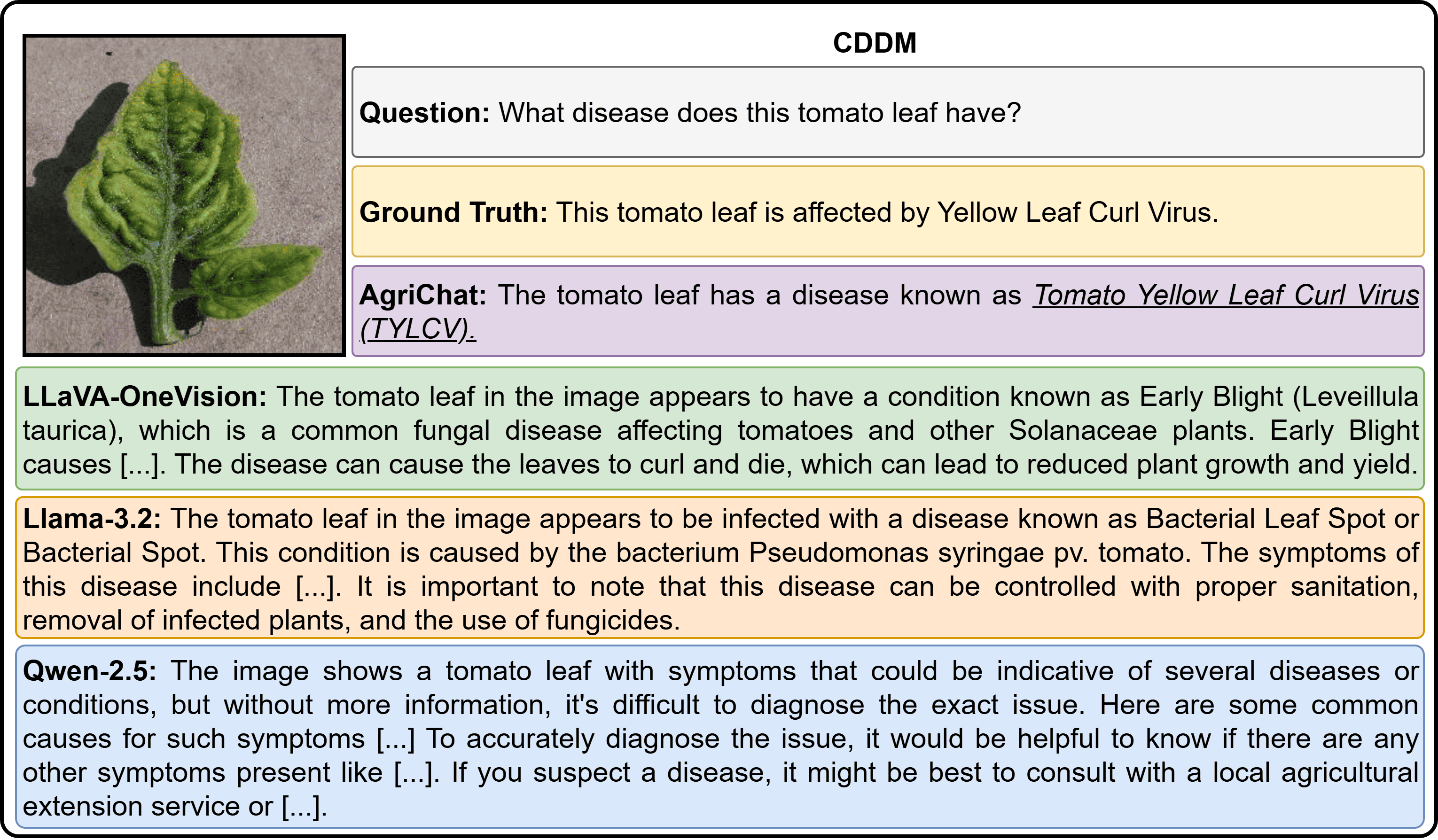}
    \caption{Qualitative comparison on a CDDM tomato disease diagnosis task.}
    \label{fig:qual5}
\end{figure}

As illustrated in Figure~\ref{fig:qual5}, AgriChat correctly identifies Tomato Yellow Leaf Curl Virus (TYLCV) in this zero-shot scenario, matching the ground truth precisely. The generalist baselines each propose different, incorrect diagnoses: LLaVA-OneVision names Early Blight, Llama-3.2 suggests Bacterial Leaf Spot, and Qwen-2.5 declines to commit to any diagnosis at all. The divergence in the baselines' outputs, confidently wrong or deliberately evasive, illustrates the lack of reliable plant pathology knowledge in generalist models. AgriChat's correct identification, achieved without any CDDM training data, suggests that agricultural fine-tuning cultivates transferable visual–semantic associations for disease recognition that extend well beyond the training distribution.
\subsection{Quantitative Results}
\label{subsec:quantitative}

Given that comparable domain-specific models (e.g., Agri-LLaVA, AgroGPT) are not publicly released as of this writing, we compared AgriChat (7B) against state-of-the-art generalist multimodal models: LLaVA-OneVision (7B), Llama-3.2 (Vision-11B), and Qwen-2.5 (VL-7B).

\subsubsection{Benchmark Performance}
\begin{figure}[htbp]

    \centering
    \includegraphics[width=\textwidth]{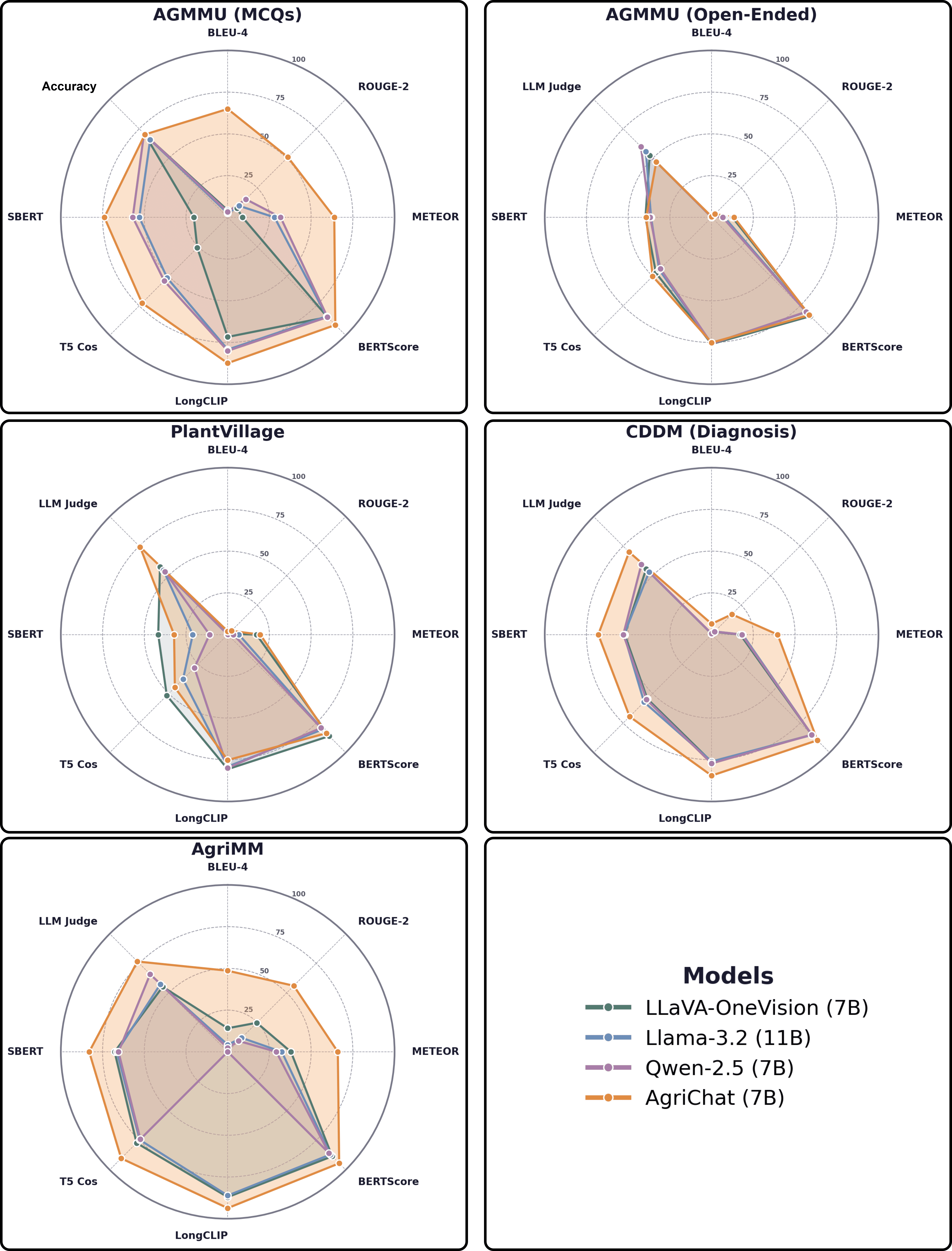} 
    \caption{Radar chart evaluation of AgriChat against baselines. AgriChat shows distinct advantages in diagnostic tasks (AgriMM, CDDM, PlantVillage) while generalist models remain competitive in open-ended reasoning.}
\label{fig:comprehensive_radar_charts}
\end{figure}

Table~\ref{tab:comprehensive_benchmark} and Figure~\ref{fig:comprehensive_radar_charts} summarize the results. AgriChat demonstrates dominant performance on the in-domain AgriMM dataset (LLM Judge: 77.43\%) and strong zero-shot transfer to PlantVillage (74.26\%) and CDDM Diagnosis (69.94\%). 

\definecolor{highlight}{RGB}{255,235,210}   


{
\newcommand{\gtext}[1]{\textcolor{teal}{\textbf{#1}\,$\uparrow$}}
\newcommand{\rtext}[1]{\textcolor{red}{\textbf{#1}\,$\downarrow$}}
\newcommand{\best}[1]{\textbf{#1}}
\newcommand{\second}[1]{\underline{#1}}

\tiny
\renewcommand{\arraystretch}{1.1}
\setlength{\tabcolsep}{3.5pt}

\setlength{\LTleft}{0pt plus 1fill}
\setlength{\LTright}{0pt plus 1fill}

\begin{longtable}{
  l                        
  l                        
  c                        
  c                        
  c                        
  >{\columncolor{highlight}}c  
  c                        
}

\caption{Comprehensive benchmark on multiple datasets. All metrics are scaled to 0--100 for clarity. \textbf{Bold} indicates the best performance; \underline{underline} indicates the second best.
\textbf{Diff} reports the absolute difference in percentage points between AgriChat and the best-performing baseline.}
\label{tab:comprehensive_benchmark} \\

\toprule
\multirow{2}{*}{\textbf{Dataset}}
  & \multirow{2}{*}{\textbf{Metric}}
  & \textbf{LLaVA-OneVision}
  & \textbf{Llama-3.2}
  & \textbf{Qwen-2.5}
  & \textbf{AgriChat}
  & \textbf{Diff} \\
  &
  & \textit{(7B)}
  & \textit{(11B)}
  & \textit{(7B)}
  & \textit{(7B)}
  & \textit{(vs Best)} \\
\midrule[\heavyrulewidth]
\endfirsthead

\toprule
\multirow{2}{*}{\textbf{Dataset}}
  & \multirow{2}{*}{\textbf{Metric}}
  & \textbf{LLaVA-OneVision}
  & \textbf{Llama-3.2}
  & \textbf{Qwen-2.5}
  & \textbf{AgriChat}
  & \textbf{Diff} \\
  &
  & \textit{(7B)}
  & \textit{(11B)}
  & \textit{(7B)}
  & \textit{(7B)}
  & \textit{(vs Best)} \\
\midrule[\heavyrulewidth]
\endhead

\midrule
\multicolumn{7}{r}{\textit{Continued on next page\,\ldots}} \\
\endfoot

\bottomrule
\endlastfoot

\multirow{11}{*}{\shortstack[l]{AGMMU\\(MCQs)}}
  & \textbf{Lexical / Math Metrics} & & & & & \\
  & \quad BLEU-4      & \second{4.20}  & 2.49  & 3.29  & \best{64.94} & \gtext{+60.7} \\
  & \quad ROUGE-2     & 7.81           & 9.73  & \second{15.36} & \best{50.98} & \gtext{+35.6} \\
  & \quad METEOR      & 8.88           & 28.06 & \second{31.70} & \best{63.87} & \gtext{+32.2} \\
\cmidrule(lr){2-7}
  & \textbf{Semantic Metrics}       & & & & & \\
  & \quad BERTScore   & \second{84.57} & 83.89 & 84.53 & \best{91.22} & \gtext{+6.7}  \\
  & \quad LongCLIP    & 71.66          & 79.24 & \second{79.98} & \best{87.38} & \gtext{+7.4}  \\
  & \quad T5 Cos      & 25.81          & 51.32 & \second{53.83} & \best{72.61} & \gtext{+18.8} \\
  & \quad SBERT       & 20.25          & 53.01 & \second{56.90} & \best{73.92} & \gtext{+17.0} \\
\cmidrule(lr){2-7}
  & \textbf{Accuracy Percentage}               & & & & & \\
  & \quad Accuracy        & 70.13         & 65.73 & \best{70.94}   & \second{70.19}& \rtext{$-$0.8} \\

\midrule[\heavyrulewidth]

\multirow{11}{*}{\shortstack[l]{AGMMU\\(Open-Ended)}}
  & \textbf{Lexical / Math Metrics} & & & & & \\
  & \quad BLEU-4      & \second{0.35}  & 0.11  & 0.09  & \best{0.43}  & \gtext{+0.1}  \\
  & \quad ROUGE-2     & \second{2.60}  & 1.04  & 0.76  & \best{2.83}  & \gtext{+0.2}  \\
  & \quad METEOR      & \second{12.45} & 7.69  & 6.62  & \best{13.44} & \gtext{+1.0}  \\
\cmidrule(lr){2-7}
  & \textbf{Semantic Metrics}       & & & & & \\
  & \quad BERTScore   & \best{83.74}   & 80.24 & 80.07 & \second{82.81}& \rtext{$-$0.9} \\
  & \quad LongCLIP    & \best{75.91}   & 75.68 & \second{75.72} & 75.08 & \rtext{$-$0.8} \\
  & \quad T5 Cos      & \second{47.17} & 44.40 & 43.47 & \best{49.95} & \gtext{+2.8}  \\
  & \quad SBERT       & \best{39.65}   & 36.30 & 36.82 & \second{39.23}& \rtext{$-$0.4} \\
\cmidrule(lr){2-7}
  & \textbf{LLM-based}              & & & & & \\
  & \quad LLM Judge (\%)            & 52.49        & \second{55.70} & \best{59.93} & 46.93 & \rtext{$-$13.0} \\

\midrule[\heavyrulewidth]

\multirow{11}{*}{\shortstack[l]{PlantVillage-\\VQA}}
  & \textbf{Lexical / Math Metrics} & & & & & \\
  & \quad BLEU-4      & \second{0.14}  & 0.08  & 0.03  & \best{2.00}  & \gtext{+1.9}  \\
  & \quad ROUGE-2     & \second{0.65}  & 0.50  & 0.22  & \best{3.18}  & \gtext{+2.5}  \\
  & \quad METEOR      & \second{17.25} & 6.72  & 3.43  & \best{19.52} & \gtext{+2.3}  \\
\cmidrule(lr){2-7}
  & \textbf{Semantic Metrics}       & & & & & \\
  & \quad BERTScore   & \best{86.02}   & \second{80.37} & 79.12 & 83.58 & \rtext{$-$2.4} \\
  & \quad LongCLIP    & \best{80.65}   & 79.11 & \second{79.89} & 75.20 & \rtext{$-$5.5} \\
  & \quad T5 Cos      & \best{51.65}   & \second{37.85} & 28.20 & 44.80 & \rtext{$-$6.9} \\
  & \quad SBERT       & \best{41.64}   & \second{20.95} & 10.90 & 32.13 & \rtext{$-$9.5} \\
\cmidrule(lr){2-7}
  & \textbf{LLM-based}              & & & & & \\
  & \quad LLM Judge (\%)            & \second{57.41}& 54.44 & 53.21 & \best{74.26} & \gtext{+16.9} \\

\midrule[\heavyrulewidth]

\multirow{11}{*}{\shortstack[l]{CDDM\\(Diagnosis)}}
  & \textbf{Lexical / Math Metrics} & & & & & \\
  & \quad BLEU-4      & 0.45           & 0.56  & \second{0.57}  & \best{6.42}  & \gtext{+5.9}  \\
  & \quad ROUGE-2     & 2.28           & \second{2.62}  & 2.52  & \best{17.16} & \gtext{+14.5} \\
  & \quad METEOR      & 17.17          & \second{18.63} & 18.11 & \best{39.59} & \gtext{+21.0} \\
\cmidrule(lr){2-7}
  & \textbf{Semantic Metrics}       & & & & & \\
  & \quad BERTScore   & 84.91          & \second{85.06} & 84.82 & \best{89.60} & \gtext{+4.5}  \\
  & \quad LongCLIP    & 76.30          & 76.03 & \second{77.21} & \best{84.50} & \gtext{+7.3}  \\
  & \quad T5 Cos      & 54.47          & \second{57.21} & 55.13 & \best{69.50} & \gtext{+12.3} \\
  & \quad SBERT       & 52.06          & 52.52 & \second{52.92} & \best{67.92} & \gtext{+15.0} \\
\cmidrule(lr){2-7}
  & \textbf{LLM-based}              & & & & & \\
  & \quad LLM Judge (\%)            & 55.53        & 53.03 & \second{59.51} & \best{69.94} & \gtext{+10.4} \\

\midrule[\heavyrulewidth]

\multirow{11}{*}{AgriMM}
  & \textbf{Lexical / Math Metrics} & & & & & \\
  & \quad BLEU-4      & \second{14.13} & 4.24  & 2.29  & \best{49.34} & \gtext{+35.2} \\
  & \quad ROUGE-2     & \second{24.61} & 11.84 & 9.24  & \best{57.29} & \gtext{+32.7} \\
  & \quad METEOR      & \second{37.89} & 32.43 & 29.11 & \best{66.70} & \gtext{+28.8} \\
\cmidrule(lr){2-7}
  & \textbf{Semantic Metrics}       & & & & & \\
  & \quad BERTScore   & \second{88.52} & 87.02 & 85.83 & \best{94.71} & \gtext{+6.2}  \\
  & \quad LongCLIP    & \second{87.05} & 86.10 & 0.00  & \best{93.97} & \gtext{+6.9}  \\
  & \quad T5 Cos      & \second{77.23} & 74.58 & 74.02 & \best{90.68} & \gtext{+13.5} \\
  & \quad SBERT       & \second{67.79} & 66.41 & 65.47 & \best{83.60} & \gtext{+15.8} \\
\cmidrule(lr){2-7}
  & \textbf{LLM-based}              & & & & & \\
  & \quad LLM Judge (\%)            & 55.12        & 57.18 & \second{65.77} & \best{77.43} & \gtext{+11.7} \\

\end{longtable}
}

\subsubsection{Inference Performance}
To assess real-world deployment viability, we benchmarked AgriChat on a consumer-grade NVIDIA RTX 3090 GPU (24GB VRAM), representative of hardware accessible to agricultural research institutions and extension services. All models were evaluated using 4-bit quantization, batch size of 1, and a standardized test image (1328 KB) with the prompt \textit{``What is the name of the plant in the image?''} repeated across 100 inference cycles.

\begin{table}[h]
\centering
\caption{Inference performance benchmarks on NVIDIA RTX 3090 (24GB VRAM). All models evaluated at 4-bit precision with batch size 1 on a 1328 KB agricultural field image. Metrics averaged over 100 repetitions. \textbf{Bold} indicates best performance; \underline{Underline} indicates second best.}
\label{tab:inference_benchmark}
\resizebox{\textwidth}{!}{%
\begin{tabular}{l c c c c c c}
\toprule
\textbf{Model} & \textbf{Load Time (s)} & \textbf{Avg Inf Time (s)} & \textbf{It/sec} & \textbf{Tokens/sec} & \textbf{Mem Load (GB)} & \textbf{Mem Peak (GB)} \\
\midrule
Llama-3.2 (Vision-11B) & 29.05 & 9.938 & 0.101 & \textbf{25.86} & \textbf{7.18} & \textbf{8.16} \\
Qwen-2.5 (VL-7B) & 22.45 & 29.769 & 0.034 & 8.60 & 11.33 & 11.76 \\
LLaVA-OneVision (7B) & \underline{9.98} & \textbf{1.546} & \textbf{0.647} & \underline{12.94} & \underline{9.99} & \underline{10.99} \\
AgriChat (7B)& \textbf{9.27} & \underline{2.315} & \underline{0.432} & 9.50 & 10.71 & 12.32 \\
\bottomrule
\end{tabular}%
}
\end{table}

As shown in Table~\ref{tab:inference_benchmark}, AgriChat achieves an average inference time of 2.315 seconds per query, enabling real-time diagnostic applications while remaining 4.3$\times$ faster than Llama-3.2-11B and 12.9$\times$ faster than Qwen-2.5-7B. The model's throughput of 0.432 iterations per second translates to approximately 1,555 diagnostic queries per hour, sufficient for field deployment via mobile applications. With a memory footprint of 10.71--12.32 GB and rapid load time of 9.27 seconds, AgriChat runs comfortably on mid-range GPUs (RTX 4060 Ti, A4000) without requiring specialized data center hardware. While the 50\% latency increase relative to base LLaVA-OneVision reflects the additional computational cost of specialized adapters, this trade-off delivers a 21.4 percentage point improvement in diagnostic accuracy on AgriMM (Table \ref{tab:comprehensive_benchmark}), making it acceptable for agricultural applications where misdiagnosis carries significant economic consequences.

\subsection{Discussion}
\label{subsec:discussion}
\subsubsection{Generalization to Unseen Domains}
A critical finding is the model's ability to generalize to datasets not seen during fine-tuning. Despite being trained exclusively on AgriMM, AgriChat achieved strong performance on CDDM (Diagnosis), leading all baselines across both lexical and semantic metrics by substantial margins (e.g., BLEU-4: +5.9\%, LLM Judge: +10.4\%). On PlantVillageVQA, AgriChat led on lexical metrics and LLM-judged quality (+16.9\%), though it trailed LLaVA-OneVision on semantic similarity metrics (BERTScore, SBERT), suggesting its outputs are structurally well-formed but phrased differently from reference answers. Together, these results suggest that training on AgriMM's diverse 63-source corpus helps prevent overfitting to specific visual conditions, enabling the model to learn robust representations of crop pathology features across different data distributions.

\subsubsection{The Specialist-Generalist Trade-off}
The results highlight an inherent trade-off. While AgriChat excels in visual diagnosis, generalist baselines retained an advantage in tasks requiring broad, open-ended knowledge retrieval, such as AGMMU (Open-Ended). AgriMM's training pipeline prioritizes ``Vision-to-Verified-Knowledge,'' creating a \textit{specialist diagnostician}. Consequently, the model is highly accurate at identification but conservative in generating treatment protocols where its training data did not provide verified advice. This behavior acts as a safety mechanism, reducing the risk of hallucinating management strategies in high-stakes agricultural contexts.

\subsubsection{Metric Sensitivity and Alignment}
The discrepancy between lexical and semantic metrics in Table~\ref{tab:comprehensive_benchmark} validates our multi-faceted evaluation approach. Generalist models achieved near-zero BLEU-4 scores on structured tasks such as AGMMU (MCQs) (e.g., LLaVA-OneVision: 0.0420, Llama-3.2: 0.0249), reflecting verbose or ill-formatted outputs, despite maintaining moderate BERTScore values. On structured diagnostic datasets (AGMMU MCQs, CDDM, AgriMM), AgriChat achieves consistently high scores across both lexical and semantic metrics, demonstrating that domain-specific fine-tuning successfully aligns both output \textit{content} and \textit{format}.

\subsection{Ablation Studies}
\label{sec:ablation}

We conduct a series of controlled ablation experiments to isolate the contribution of each design decision in our fine-tuning framework. Specifically, we examine three axes of variation: (i)~the rank of the LoRA adapter matrices, (ii)~whether the vision encoder benefits from lightweight adaptation alongside the language decoder, and (iii)~single-stage versus two-stage curriculum training. All fine-tuning runs and evaluations are carried out on our dataset AgriMM. Unless otherwise stated, each experiment varies a single factor while holding all remaining hyperparameters at their baseline values. Results are reported across three complementary evaluation dimensions: lexical overlap (BLEU-4, ROUGE-2, METEOR), semantic similarity (BERTScore, LongCLIP, T5 Cosine, SBERT), and holistic judgement quality (LLM Judge).

\subsubsection{Effect of LoRA Rank}
\label{sec:ablation_rank}

The rank $r$ of the low-rank adapter matrices governs the capacity available for domain adaptation: an insufficiently low rank risks underfitting the specialized vocabulary and reasoning patterns of agricultural advisory dialogue, whereas an excessively high rank invites overfitting given the limited size of the training corpus. We evaluate $r \in \{32, 64, 128\}$, maintaining the relationship $\alpha = 2r$ throughout so that the effective learning-rate scaling remains constant across configurations. As shown in Table~\ref{tab:ablation_lora_rank}, increasing the rank from 32 to 64 yields substantial gains across all metrics, with BLEU-4 improving by roughly 10\% relative and LLM Judge accuracy rising by over 3 percentage points. Further doubling the rank to 128 produces marginal additional improvements on the semantic metrics while leaving LLM Judge essentially unchanged (76.39\% vs.\ 76.54\%). These results suggest that $r{=}64$ captures the bulk of the adaptation benefit; nonetheless, we adopt $r{=}128$ for subsequent experiments to maximise headroom for the more challenging ablation conditions that follow.
\begin{table}[ht]
    \centering
    \footnotesize
    \renewcommand{\arraystretch}{1.0}
    \caption{Effect of LLM LoRA rank on AgriMM. \textbf{Bold} denotes the best result per metric; \underline{underline} denotes the second best.}
    \label{tab:ablation_lora_rank}
    \begin{tabular}{l c c c}
        \toprule
        \textbf{Metric} & \textbf{Rank 32} & \textbf{Rank 64} & \textbf{Rank 128} \\
        \midrule
        \multicolumn{4}{l}{\textit{Lexical}} \\
        BLEU-4    & 0.4422 & \textbf{0.4860} & \textbf{0.4860} \\
        ROUGE-2   & 0.5146 & \underline{0.5598} & \textbf{0.5663} \\
        METEOR    & 0.6172 & \textbf{0.6599} & \underline{0.6588} \\
        \midrule
        \multicolumn{4}{l}{\textit{Semantic}} \\
        BERTScore & 0.9377 & \underline{0.9447} & \textbf{0.9461} \\
        LongCLIP  & 0.9258 & \underline{0.9365} & \textbf{0.9378} \\
        T5 Cos    & 0.8850 & \underline{0.9027} & \textbf{0.9035} \\
        SBERT     & 0.8064 & \underline{0.8302} & \textbf{0.8317} \\
        \midrule
        \multicolumn{4}{l}{\textit{LLM-based}} \\
        LLM Judge (\%) & 73.46 & \textbf{76.54} & \underline{76.39} \\
        \bottomrule
    \end{tabular}
\end{table}
\subsubsection{Effect of Vision Encoder Adaptation}
\label{sec:ablation_vision}

In the default configuration the SigLIP vision encoder remains fully frozen and LoRA adapters are injected exclusively into the Qwen-2 language decoder. This design assumes that the general-purpose visual features learned during large-scale pre-training transfer adequately to the agricultural domain. To test this assumption, we introduce an additional set of rank-32 LoRA adapters into the vision encoder's linear projection layers and retrain under otherwise identical conditions (LLM rank 128, single-stage training). Table~\ref{tab:ablation_vision_lora} reports the results. Adapting the vision encoder yields consistent improvements across every metric, with the largest relative gains observed on the lexical measures (BLEU-4: $+1.52\%$; ROUGE-2: $+1.17\%$) and on the LLM Judge score ($+1.36\%$). The semantic similarity metrics also improve, albeit more modestly (BERTScore: $+0.11\%$; SBERT: $+0.52\%$). These findings indicate that domain-specific visual representations constitute a meaningful, if secondary, performance bottleneck: fine-grained agricultural features such as lesion morphology, growth-stage appearance, and plant anatomical detail benefit from lightweight encoder adaptation beyond what frozen pre-trained features provide.

{
\small
\renewcommand{\arraystretch}{1.15} 
\setlength{\tabcolsep}{3.5pt}

\newcommand{\gtext}[1]{\textcolor{teal}{\textbf{#1} $\uparrow$}}
\newcommand{\rtext}[1]{\textcolor{red}{\textbf{#1} $\downarrow$}}
\newcommand{\best}[1]{\textbf{#1}}
\newcommand{\second}[1]{\underline{#1}}
\newcommand{\sectionrow}[1]{\multicolumn{3}{l}{\textit{\textbf{#1}}}}

\begin{table}[ht]
    \centering
    \footnotesize
    \renewcommand{\arraystretch}{1.15}
    \setlength{\tabcolsep}{8pt}
    \caption{Effect of vision encoder adaptation on AgriMM. Both configurations use LLM rank 128 with single-stage training.}
    \label{tab:ablation_vision_lora}
    \begin{tabular}{l c c}
        \toprule
        \textbf{Metric} & \textbf{w/o Vision LoRA} & \textbf{w/ Vision LoRA} \\
        \midrule
        \multicolumn{3}{l}{\textit{Lexical}} \\
        BLEU-4    & \second{0.4860} & \best{0.4934} \\
        ROUGE-2   & \second{0.5663} & \best{0.5729} \\
        METEOR    & \second{0.6588} & \best{0.6670} \\
        \midrule
        \multicolumn{3}{l}{\textit{Semantic}} \\
        BERTScore & \second{0.9461} & \best{0.9471} \\
        LongCLIP  & \second{0.9378} & \best{0.9397} \\
        T5 Cos    & \second{0.9035} & \best{0.9068} \\
        SBERT     & \second{0.8317} & \best{0.8360} \\
        \midrule
        \multicolumn{3}{l}{\textit{LLM-based}} \\
        LLM Judge (\%) & \second{76.39} & \best{77.43} \\
        \bottomrule
    \end{tabular}
\end{table}
}
\subsubsection{Effect of Training Strategy}
\label{sec:ablation_stage}

Our default protocol trains the model in a single stage on the full multimodal dataset. We compare this against a two-stage curriculum motivated by the hypothesis that establishing linguistic grounding before introducing visual alignment may yield more stable optimisation. In Stage~1, LoRA adapters are applied solely to the language decoder and trained on text-only agricultural QA pairs, allowing the model to internalise domain terminology, disease nomenclature, and advisory reasoning patterns. In Stage~2, adapters are additionally introduced into the vision encoder and training proceeds on the complete multimodal corpus. Contrary to expectation, the two-stage curriculum substantially underperforms the single-stage baseline across all evaluation axes (Table~\ref{tab:ablation_training_stage}). The degradation is particularly pronounced on lexical metrics, where BLEU-4 drops by 15.12\% relative and ROUGE-2 by 16.32\%. Semantic scores and the LLM Judge similarly decline. We attribute this to catastrophic forgetting during the stage transition: the adapter weights optimised for text-only supervision in Stage~1 provide a suboptimal initialisation for the multimodal objective in Stage~2, effectively undoing a portion of the linguistic adaptation. These results firmly justify the simpler single-stage protocol, which benefits from joint vision--language optimisation from the outset.
{
\small
\renewcommand{\arraystretch}{1.15} 
\setlength{\tabcolsep}{3.5pt}

\newcommand{\best}[1]{\textbf{#1}}
\newcommand{\second}[1]{\underline{#1}}
\newcommand{\sectionrow}[1]{\multicolumn{3}{l}{\textit{\textbf{#1}}}}

\begin{table}[ht]
    \centering
    \footnotesize
    \renewcommand{\arraystretch}{1.15}
    \setlength{\tabcolsep}{8pt}
    \caption{Effect of training strategy on AgriMM. Both configurations use LLM rank 128 and vision rank 32.}
    \label{tab:ablation_training_stage}
    \begin{tabular}{l c c}
        \toprule
        \textbf{Metric} & \textbf{Single-stage} & \textbf{Two-stage} \\
        \midrule
        \multicolumn{3}{l}{\textit{Lexical}} \\
        BLEU-4    & \best{0.4934} & \second{0.4286} \\
        ROUGE-2   & \best{0.5729} & \second{0.4925} \\
        METEOR    & \best{0.6670} & \second{0.6103} \\
        \midrule
        \multicolumn{3}{l}{\textit{Semantic}} \\
        BERTScore & \best{0.9471} & \second{0.9364} \\
        LongCLIP  & \best{0.9397} & \second{0.9274} \\
        T5 Cos    & \best{0.9068} & \second{0.8869} \\
        SBERT     & \best{0.8360} & \second{0.8057} \\
        \midrule
        \multicolumn{3}{l}{\textit{LLM-based}} \\
        LLM Judge (\%) & \best{77.43} & \second{73.26} \\
        \bottomrule
    \end{tabular}
\end{table}
}

\section{Conclusion}
\label{sec:conclusion}

This work addressed two fundamental bottlenecks limiting the deployment of MLLMs in agriculture: the scarcity of scientifically verified training data and the lack of domain-specialized models capable of fine-grained agricultural reasoning. To overcome the first, we introduced the \textit{Vision-to-Verified-Knowledge (V2VK)} pipeline, which synthesizes high-quality VQA pairs by grounding the outputs of multiple generative models in verified phytopathological literature. This produced AgriMM, a publicly available benchmark of 121,425 images and 607,125 VQA pairs spanning more than 3,000 agricultural classes across 63 source datasets, validated through a rigorous human-in-the-loop protocol. To overcome the second, we presented AgriChat, a domain-specialized MLLM built on LLaVA-OneVision and adapted through parameter-efficient fine-tuning of both the vision encoder and language decoder. AgriChat achieves state-of-the-art in-domain performance in both internal and external benchmarks and strong zero-shot generalization, outperforming open-source models, while working on consumer-grade hardware. Future work will focus on three directions: expanding AgriMM to cover pest identification, increasing per-class image representation to support more robust learning, and incorporating more recent generative models into both the V2VK pipeline and the AgriChat architecture. All data, code, and model weights are publicly released to support reproducibility and to serve as a foundation for future research in agricultural multimodal intelligence.
\appendix
\onecolumn

\section{Dataset Composition}
\label{app:datasets}

This section details the specific source datasets aggregated to construct AgriMM. To ensure reproducibility, we list the exact dataset identifiers used in our aggregation pipeline.

\subsection{Crop Counting and Spatial Reasoning Sources}
\label{app:detection_datasets}
The 33 object detection datasets used to construct the quantitative reasoning component of AgriMM are listed below. These datasets were processed to generate ground-truth counts for instruction tuning.

\begin{itemize}[leftmargin=*]
    \setlength\itemsep{0em}
    \item \textbf{Fruit \& Nut Detection:} almond\_bloom\_2023 \cite{agml}, almond\_harvest\_2021 \cite{agml}, apple\_detection\_drone\_brazil \cite{agml}, apple\_detection\_spain \cite{agml}, apple\_detection\_usa \cite{agml}, embrapa\_wgisd\_grape\_detection \cite{agml}, fruit\_detection\_worldwide \cite{agml}, grape\_detection\_californiaday \cite{agml}, grape\_detection\_californianight \cite{agml}, grape\_detection\_syntheticday \cite{agml}, mango\_detection\_australia \cite{agml}, Orange\_dataset, strawberry\_detection\_2022 \cite{agml}, strawberry\_detection\_2023 \cite{agml}, tomato\_ripeness\_detection \cite{agml}, wGrapeUNIPD-DL \cite{sozzi_marco_2022_4066730}.
    \item \textbf{Field Crops \& Vegetables:} ghai\_broccoli\_detection \cite{agml}, ghai\_green\_cabbage\_detection \cite{agml}, ghai\_iceberg\_lettuce\_detection \cite{agml}, ghai\_romaine\_detection \cite{agml}, GWHD2021 (Global Wheat Head Detection) \cite{DAVID20219846158}, wheat\_head\_counting \cite{agml}.
    \item \textbf{Specialized \& General Agriculture:} CBDA \cite{10287390, YE20231004}, DRPD \cite{teng2023paniclecloud}, gemini\_flower\_detection\_2022 \cite{agml}, gemini\_leaf\_detection\_2022 \cite{agml}, gemini\_plant\_detection\_2022 \cite{agml}, gemini\_pod\_detection\_2022 \cite{agml}, MTDC \cite{zou2020maize}, plant\_doc\_detection \cite{agml}, SHC \cite{SHC}, WEDD \cite{madec2019ear}, YOLOPOD \cite{xiang2023yolo}.
\end{itemize}

\subsection{Disease and Stress Classification Sources}
\label{app:classification_datasets}
The 29 classification datasets comprising the pathological component of the benchmark are listed below. Each dataset includes a ``Healthy'' baseline class alongside specific biotic and abiotic stress classes.
arabica\_coffee\_leaf\_disease\_classification \cite{agml}, banana\_leaf\_disease\_classification \cite{agml}, bean\_disease\_uganda \cite{agml}, betel\_leaf\_disease\_classification \cite{agml}, blackgram\_plant\_leaf\_disease\_classification \cite{agml}, chilli\_leaf\_classification \cite{agml}, coconut\_tree\_disease\_classification \cite{agml}, corn\_maize\_leaf\_disease \cite{agml}, crop\_weeds\_greece \cite{agml}, cucumber\_disease\_classification \cite{agml}, guava\_disease\_pakistan \cite{agml}, java\_plum\_leaf\_disease\_classification \cite{agml}, leaf\_counting\_denmark \cite{agml}, onion\_leaf\_classification \cite{agml}, orange\_leaf\_disease\_classification \cite{agml}, paddy\_disease\_classification \cite{agml}, papaya\_leaf\_disease\_classification \cite{agml}, plant\_doc\_classification \cite{agml}, plant\_seedlings\_aarhus \cite{agml}, plant\_village\_classification \cite{agml}, rangeland\_weeds\_australia \cite{agml}, rice\_leaf\_disease\_classification \cite{agml}, riseholme\_strawberry\_classification\_2021 \cite{agml}, soybean\_weed\_uav\_brazil \cite{agml}, sugarcane\_damage\_usa \cite{agml}, sunflower\_disease\_classification \cite{agml}, tea\_leaf\_disease\_classification \cite{agml}, tomato\_leaf\_disease \cite{agml}, vine\_virus\_photo\_dataset \cite{agml}.

\section{Synthesis Pipeline Prompt Templates}
\label{app:prompts}

We provide the exact system prompts used in our \textbf{Vision-to-Verified-Knowledge} pipeline. These prompts were designed to minimize hallucination by enforcing strict constraints on the language models.

\subsection{Stage 1: Visual Grounding (Image Captioning)}
\label{app:stage1_prompt}
\textbf{Model:} Gemma 3 (12B) \\
\textbf{Purpose:} To generate structured visual metadata rather than generic captions.

\begin{small}
\begin{verbatim}
Write a descriptive caption of about 3-5 sentences given that 
the image contains {extra_details}.
Include these aspects if clearly visible:
- Crop name and type
- Growth stage (seedling/vegetative/flowering/fruiting/harvest)
- Ground cover and plant density
- Image perspective (top-down/oblique/side/macro/unknown)
- Environmental conditions (field/greenhouse/laboratory)
- Plant health indicators
Rules for caption:
- Use clear, neutral language
- No speculation - only describe what is visible
- If something cannot be determined, use 'unknown'
- Write as natural sentences
\end{verbatim}
\end{small}

\subsection{Stage 2: Knowledge Retrieval}
\label{app:stage2_prompt}
\textbf{Model:} Gemini 3 Pro (Web Search Enabled) \\
\textbf{Purpose:} To retrieve verified botanical and phytopathological knowledge.

\subsubsection{Species Identification Prompt}
\begin{small}
\begin{verbatim}
For the class name {class_names}, generate a 
detailed botanical description paragraph (~300 words) covering:
- Taxonomic classification (family, genus, species)
- Morphological characteristics (leaf shape, stem structure, 
  inflorescence, fruit morphology)
- Native habitat and biogeographic distribution
- Cultivation requirements (soil, climate, water)
- Ecological significance and agricultural use
Format your output as {"class_name":"detailed discription"}
\end{verbatim}
\end{small}

\subsubsection{Disease Classification Prompt}
\begin{small}
\begin{verbatim}
For each class name in {disease_class_names}, generate a 
detailed paragraph (~300 words) providing an integrated account of:
- Plant taxonomy, morphology, and natural habitat
- Disease etiology (causal agent, pathogen taxonomy)
- Visible symptoms (lesion morphology, discoloration patterns, 
  necrosis, wilting)
- Affected plant organs (leaves, stems, fruits, roots)
- Pathogen biology and infection cycle
- Environmental factors influencing disease development
- Comparison with healthy plant phenotype
When the class represents "Healthy", describe the ideal 
botanical state emphasizing vigor, normal morphology, and 
optimal appearance.
\end{verbatim}
\end{small}

\subsection{Stage 3: Instruction Generation}
\label{app:stage3_prompt}
\textbf{Model:} LLaMA 3.1 8B Instruct \\
\textbf{Purpose:} To synthesize visual captions and retrieved knowledge into QA pairs.

\begin{small}
\begin{verbatim}
You are an expert agricultural AI trainer. Generate exactly 
5 high-quality, diverse QA pairs.

**SOURCE DATA:**
- Additional Info: {class_info}
- Image Caption: {caption}

**STRICT RULES:**
1. GROUNDING: Use ONLY provided info. If the image/info 
   doesn't mention a disease, don't invent one.
2. FORMAT: Output a single JSON array of 5 objects.
3. ANSWER STYLE: Use full, professional sentences. 
   Instead of "okra," say "The image shows an okra plant 
   (Abelmoschus esculentus)."

**REQUIRED QUESTION CATEGORIES (One per slot):**
1. Identification: Identify the plant and its variety.
2. Visual Reasoning: Ask HOW the plant can be identified 
   (e.g., "What visual features distinguish this species?").
3. Condition & Health: Ask about leaf/fruit/stem state 
   (color, spots, growth stage).
4. Cultivation Knowledge: Connect visuals to agronomic 
   requirements (e.g., "What are this plant's soil pH needs?").
5. Anatomy/Detail: Ask about a specific visible part 
   (flower, fruit, leaf structure).
\end{verbatim}
\end{small}

\section{Evaluation Prompts}
\label{app:eval_prompts}

\subsection{LLM-as-a-Judge Prompt}
\textbf{Model:} Qwen3-30B-A3B-Instruct-2507 \\
\textbf{Purpose:} To evaluate response quality while penalizing verbosity and hallucination.

\begin{small}
\begin{verbatim}
You are an expert evaluator assessing an AI model's response. 
Evaluate systematically and objectively.

**QUESTION**: {question}

**GROUND TRUTH (Correct Answer)**: 
{ground_truth}

**MODEL OUTPUT (To Evaluate)**:
{model_output}

---

**EVALUATION CRITERIA**
1. **Correctness**: Does the output contain the correct information from the 
Ground Truth?
2. **Completeness**: Does the output include all important information from 
Ground Truth?
3. **Clarity**: Is the output clear, well-organized, and easy to understand?
4. **Conciseness**: Is the output appropriately concise without unnecessary content?

---

**SCORING RUBRIC** (1-4 scale)
**Score 1 (Poor)**: Major deficiencies. Factually incorrect or missing multiple 
key facts.
**Score 2 (Fair)**: Significant issues. Missing 1-2 important facts or minor 
inaccuracies.
**Score 3 (Good)**: Solid with minor issues only. Factually accurate but maybe 
slightly verbose 
or misses tiny details.
**Score 4 (Excellent)**: Outstanding quality. Perfectly accurate, complete, clear, 
and concise.

---

**OUTPUT**: Provide evaluation in this EXACT JSON format:
{{
  "score": <integer 1-4>,
  "justification": "1-2 sentence summary"
}}

Begin evaluation:
\end{verbatim}
\end{small}

\bibliographystyle{elsarticle-num}
\bibliography{references2}

\end{document}